\documentclass[11pt]{article}

\usepackage[final]{acl}

\usepackage{microtype}

\usepackage{graphicx}
\usepackage{booktabs}
\usepackage{makecell}
\usepackage{pfnote}
\usepackage{multirow}

\usepackage{titlesec}
\titlespacing*{\paragraph}{0pt}{0.5ex}{0.5em}

\usepackage{enumitem}
\setlist[itemize]{
	leftmargin=1.5em,
	itemsep=0pt,
	topsep=2pt,
	parsep=0pt,
	partopsep=0pt
}
\setlist[enumerate]{
	leftmargin=1.5em,
	itemsep=0pt,
	topsep=2pt,
	parsep=0pt,
	partopsep=0pt
}

\usepackage{footmisc}
\usepackage{balance}

\usepackage{stfloats}

\usepackage[english,bidi=default]{babel} 
\babelfont{rm}{TeXGyreTermesX} 
\babelprovide[import]{chinese}
\babelfont[chinese]{rm}{NotoSerifSC.ttf}

\title{\raisebox{-1.65ex}{\protect\includegraphics[height=3.3\fontcharht\font`\0]{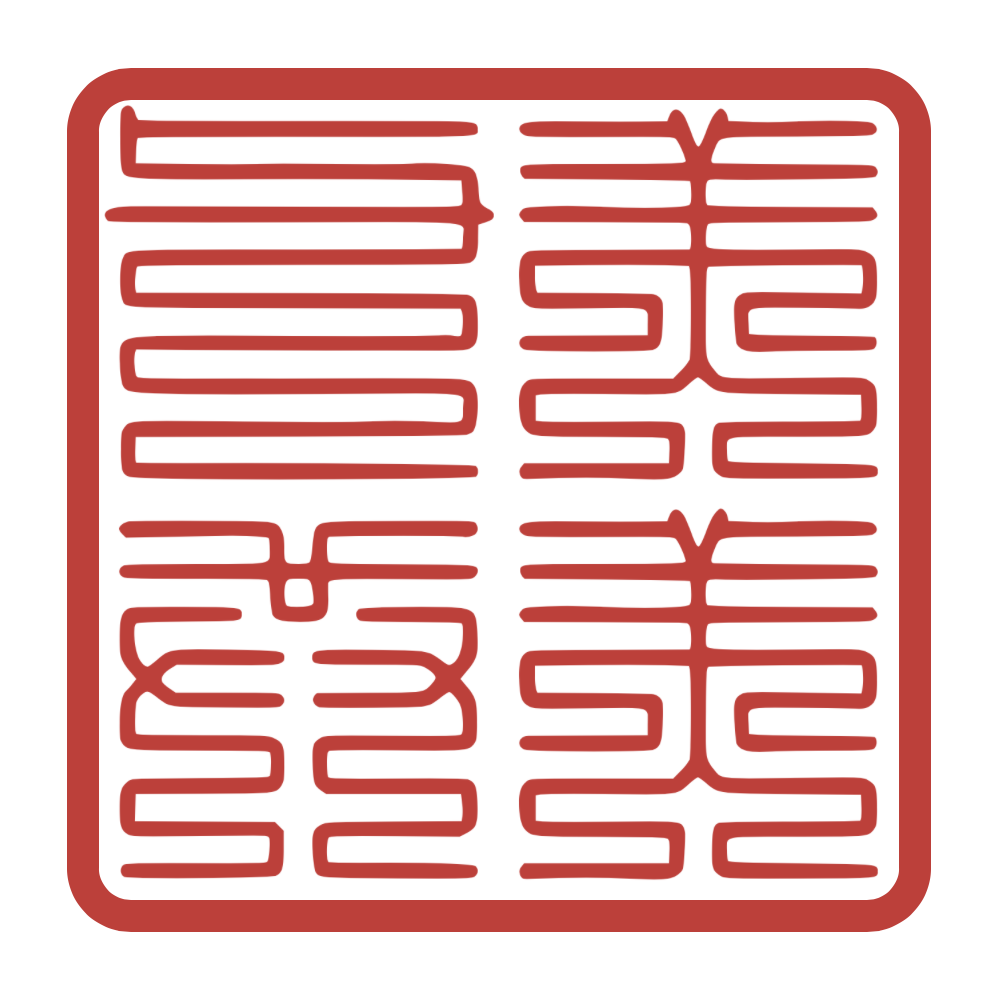}} Diversity in Unity, Theory in Practice:\\Hierarchical Multitask Benchmarks for Chinese Minority Languages}

\makeatletter
\def\thanks#1{\protected@xdef\@thanks{\@thanks\protect\footnotetext{#1}}}
\makeatother

\author{
	\textbf{Yijie Li$^{\spadesuit\diamondsuit\dagger}$}\textbf{,}~
	\textbf{Xi Cao$^{\heartsuit\clubsuit\diamondsuit\dagger}$}\textbf{,}~
	\textbf{Yuan Sun$^{\spadesuit\clubsuit\diamondsuit\ast}$}\textbf{,}~\\
	\textbf{Quulgan Minggad$^{\spadesuit\diamondsuit}$}\textbf{,}~
	\textbf{Abdulla Ablikim$^{\spadesuit\diamondsuit}$}\textbf{,}~
	\textbf{JiaQingCaiWang$^{\spadesuit\diamondsuit}$}\\
	$^{\spadesuit}$School of Information Engineering, Minzu University of China\\
	$^{\heartsuit}$School of Philosophy and Religious Studies, Minzu University of China\\
	$^{\clubsuit}$Institute of National Security, Minzu University of China\\
	$^{\diamondsuit}$National Language Resource Monitoring \& Research Center | Minority Languages Branch\\
	\texttt{liyijie@muc.edu.cn},
	\texttt{caoxi@muc.edu.cn},
	\texttt{sunyuan@muc.edu.cn}
	\thanks{$^{\dagger}$~Equal contribution.}
	\thanks{$^{\ast}$~Corresponding author.}
}

\begin{document}

\maketitle
\begin{abstract}
Despite the rapid advancement of LLMs, their performance on linguistically and culturally diverse minority languages within a unified national context remains underexplored.
We present \textsc{CMiLBench}, a collection of hierarchical multitask benchmarks designed to translate theoretical notions of \textit{diversity in unity} (in Chinese: ``\foreignlanguage{Chinese}{美美与共}'') into practical evaluation for three representative Chinese minority languages: Tibetan, Mongolian, and Uyghur.
\textsc{CMiLBench} comprises 24,663 instances across 5 difficulty levels and 17 tasks spanning foundational ability, cultural specificity, and safety alignment.
We adopt existing dataset adaptation, minority knowledge construction, and high-resource benchmark translation to construct \textsc{CMiLBench}.
We assess 14 state-of-the-art commercial and open-source LLMs with a hybrid framework that integrates automatic metrics and LLM-as-a-Judge scoring.
The comparative experimental results reveal the gap between theoretical capability and practical utility.
\textsc{CMiLBench} serves as a foundational and scalable evaluation resource to bridge the digital language divide and promote the informatization and intelligentization of low-resource Chinese minority languages.
More information about \textsc{CMiLBench} is available at our project page: \texttt{\url{https://github.com/CMLI-NLP/CMiLBench}}.
\end{abstract}

\section{Introduction}

\begin{figure}[t]
	\centering
	\includegraphics[width=0.86\linewidth]{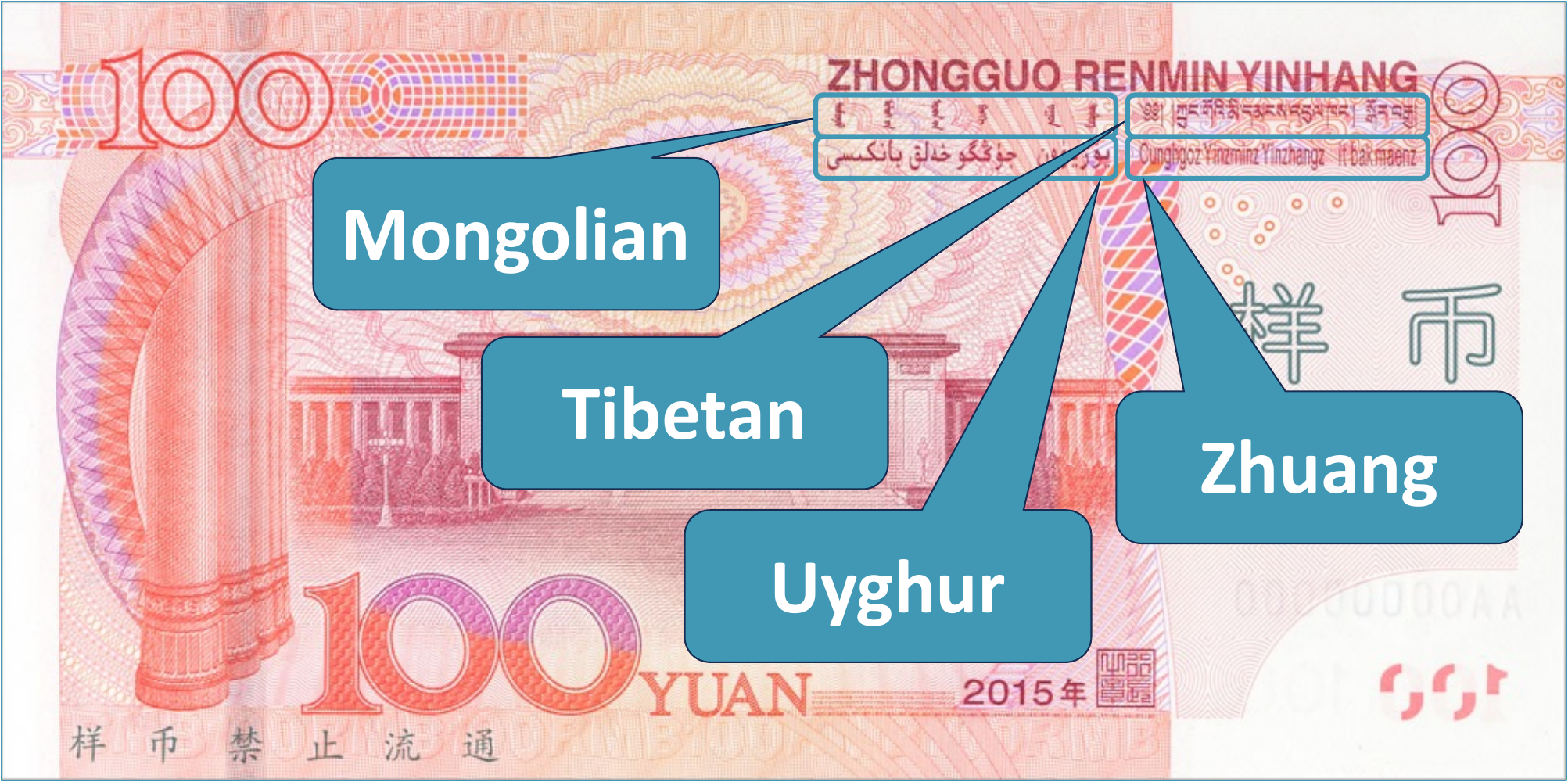}
	\captionsetup{justification=justified, singlelinecheck=false}
	\caption{Mongolian, Tibetan, Uyghur, and Zhuang on RMB. They all mean \textit{The People's Bank of China}. These four languages span different language families, diverse grammatical typologies, and distinct writing systems.}
	\label{RMB}
\end{figure}

\begin{figure}[h!]
	\centering
	\includegraphics[width=\linewidth]{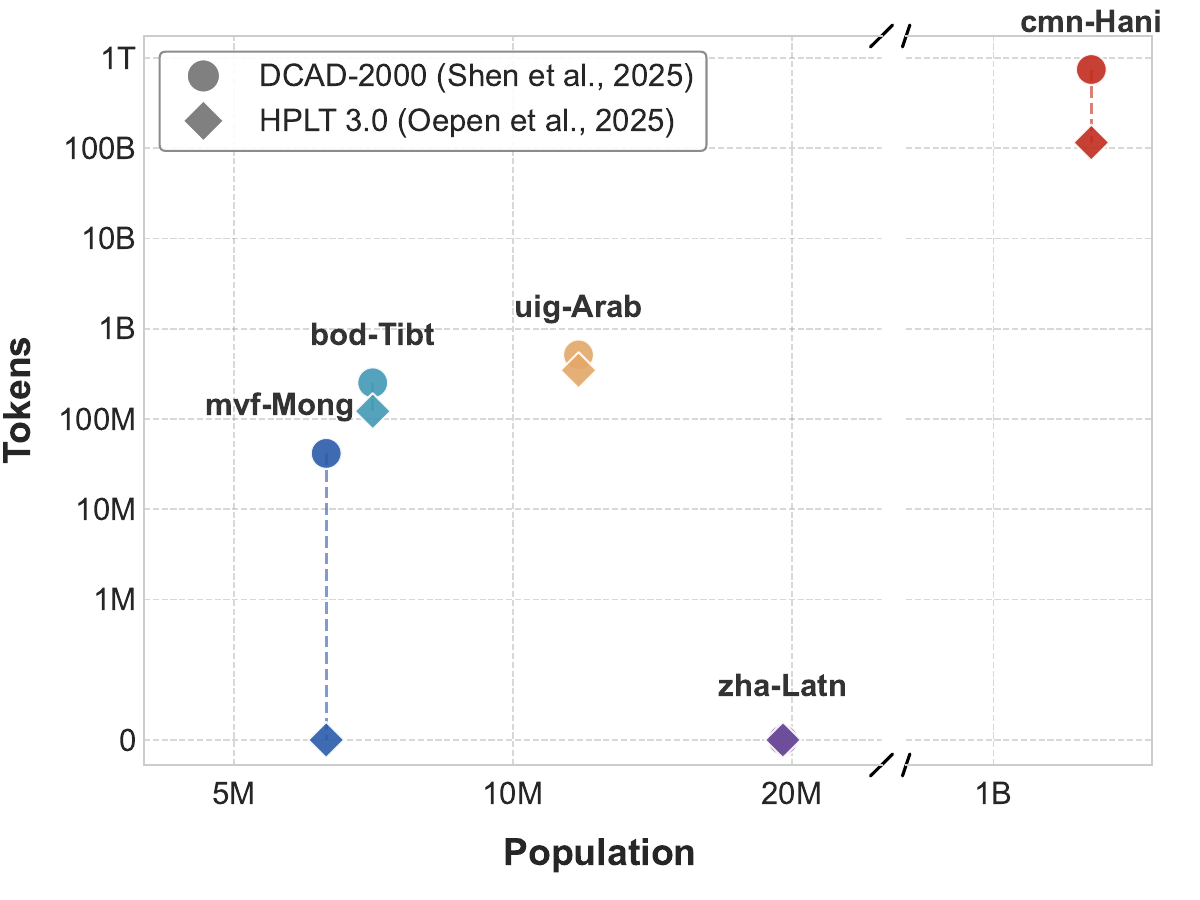}
	\captionsetup{justification=justified, singlelinecheck=false}
	\caption{The population of native speakers vs. the number of tokens in DCAD-2000 \citep{DBLP:journals/corr/abs-2502-11546} and HPLT 3.0 \citep{DBLP:journals/corr/abs-2511-01066}, two newest large-scale multilingual corpora. \texttt{mvf-Mong}, \texttt{bod-Tibt}, \texttt{uig-Arab}, \texttt{zha-Latn}, and \texttt{cmn-Hani} follow the naming convention of \texttt{\{ISO 639-3 Code\}-\{ISO 15924 Code\}}. We obtain the population data from \textit{China Statistical Yearbook 2025}.}
	\label{population_tokens}
\end{figure}


\begin{table*}[t]
	\small
	\centering
	\begin{tabular}{ccccc}
		\toprule
		\textbf{Name} & \textbf{ISO 639-3} & \textbf{ISO 15924} & \textbf{Language Family} & \textbf{Writing System}  \\
		\midrule
		Standard Chinese & \texttt{cmn} & \texttt{Hani} & Sino-Tibetan & Chinese Characters \\
		Tibetan & \texttt{bod} & \texttt{Tibt} & 	
		Sino-Tibetan & Tibetan Script \\
		Mongolian & \texttt{mvf} & \texttt{Mong} & Mongolic & Traditional Mongolian Script \\
		Uyghur & \texttt{uig} & \texttt{Arab} & Turkic & Uyghur Arabic Script \\
		Zhuang & \texttt{zha} & \texttt{Latn} & Kra-Dai & Zhuang Latin Script \\
		\bottomrule
	\end{tabular}
	\caption{ISO 639-3 codes, ISO 15924 codes, languages families, and writing systems of Standard Chinese and Chinese minority languages: Tibetan, Mongolian, Uyghur, and Zhuang.}
	\label{language_details}
\end{table*}

The rapid evolution of LLMs has revolutionized NLP, demonstrating exceptional capabilities in understanding, generation, and reasoning. However, this progress is unevenly distributed. While high-resource languages like English and Chinese benefit from massive corpora, minority languages often face a severe digital language divide. This disparity limits the accessibility of intelligent technologies for millions of native speakers and hinders the preservation of linguistic diversity in the digital age.

In the context of China, a unified multi-ethnic nation, the notion of \textit{diversity in unity} (in Chinese: ``\foreignlanguage{Chinese}{美美与共}'') is central to social development. As illustrated in Figure \ref{RMB}, Mongolian, Tibetan, Uyghur, and Zhuang appear alongside Chinese Pinyin on the Renminbi (RMB), symbolizing their linguistical and cultural significance. Despite their importance, these languages suffer from an extreme scarcity of digital resources. As shown in Figure \ref{population_tokens}, in the two newest large-scale multilingual corpora, DCAD-2000 \citep{DBLP:journals/corr/abs-2502-11546} and HPLT 3.0 \citep{DBLP:journals/corr/abs-2511-01066}, these languages (\texttt{mvf-Mong}, \texttt{bod-Tibt}, \texttt{uig-Arab}, and \texttt{zha-Latn}) are in a significantly resource-poor state compared to Standard Chinese (\texttt{cmn-Hani}). Our current study does not cover Zhuang, as even the two newest large-scale multilingual corpora do not include this language and high-quality data is difficult to acquire.

Existing evaluations for Chinese minority languages primarily focus on foundational NLP tasks \citep{yang-etal-2022-cino, DBLP:conf/smc/DengSYB0Z23, zhang-etal-2025-milic}. While valuable, these datasets often lack the complexity required to assess the advanced cognitive capabilities of modern LLMs, such as cultural perception, domain-specific knowledge, and safety alignment. Furthermore, benchmarks constructed solely via translation from high-resource languages, e.g., translating GLUE \citep{wang-etal-2018-glue} or MMLU \citep{DBLP:conf/iclr/HendrycksBBZMSS21}, fail to capture the unique ethnolinguistic characteristics and indigenous knowledge of the target languages.

To bridge the gap between theoretical capability and practical utility, we present \textsc{CMiLBench}, a collection of hierarchical multitask benchmarks designed for three representative Chinese minority languages: Tibetan, Mongolian, and Uyghur. \textsc{CMiLBench} distinguishes itself through a systematic construction pipeline that balances foundational ability, cultural specificity, and safety alignment. By integrating data from three distinct sources --- existing datasets, minority-knowledge-centric materials, and high-resource benchmarks --- we ensure diversity coverage, cultural authenticity, and data quality.

Our evaluation of 14 state-of-the-art commercial and open-source LLMs reveals significant performance disparities. Furthermore, we introduce a rigorous difficulty stratification mechanism, categorizing instances into five levels to facilitate fine-grained model diagnosis.

In summary, our contributions are as follows:

(1) We propose \textsc{CMiLBench}, a collection of hierarchical multitask benchmarks comprising 24,663 instances across 17 tasks and 5 difficulty levels, specifically tailored for Tibetan, Mongolian, and Uyghur within a unified national context.

(2) We implement a four-stage construction pipeline that processes minority-knowledge-centric materials, effectively supplementing the lack of cultural depth in translation-based benchmarks.

(3) We conduct an extensive evaluation of 14 mainstream LLMs using a hybrid framework of automatic metrics and LLM-as-a-Judge scoring. Our analysis uncovers the gap between theoretical capability and practical utility, serving as a roadmap for future development in low-resource minority languages.

\section{Related Work}

\begin{table*}[t]
	\centering
	\small
	\renewcommand{\arraystretch}{1.3}
	\setlength{\tabcolsep}{5pt}
	\begin{tabular}{p{0.11\textwidth} p{0.31\textwidth} p{0.24\textwidth} p{0.24\textwidth}}
		\toprule
		\textbf{Feature} & \textbf{\textsc{CMiLBench} (Ours)} & \makecell[l]{\textbf{MiLiC-Eval}\\\textbf{\citep{zhang-etal-2025-milic}}} & \makecell[l]{\textbf{TLUE}\\\textbf{\citep{gao-etal-2025-tlue}}} \\
		\midrule
		\textbf{Size} & 
		\textbf{24,663} instances & 
		$\approx$ 24,000 instances & 
		22,963 instances \\
		
		\textbf{Language} & 
		3 (Tibetan, Mongolian, Uyghur) & 
		4 (Tibetan, Uyghur, Kazakh, Mongolian) & 
		1 (Tibetan) \\
		
		\textbf{Data Source} & 
		\textbf{Hybrid}: \textbf{Minority-knowledge-centric Materials (32\%)}, Existing Datasets, \& High-resource Benchmarks & 
		\textbf{Adaptation}: Mainly derived or translated from existing datasets. & 
		\textbf{Translation}: Fully translated from CMMLU \& SafetyBench. \\
		
		\textbf{Cultural Specificity} & 
		\textbf{High}: Includes culture- and domain-specific QA generated from minority-knowledge-centric content (history, folklore, traditional medicine, etc.). & 
		\textbf{Limited}: Focuses on linguistic parallelism rather than cultural depth. & 
		\textbf{Limited}: Relies on benchmark translation rather than indigenous culture. \\
		
		\textbf{Task Scope} & 
		\textbf{17 Tasks}: General NLP, Specific Knowledge, \& Safety Alignment & 
		\textbf{9 Tasks}: Foundational Tasks (Vocabulary Understanding, Topic Classification, Reading Comprehension, etc.) & 
		\textbf{Inherited Tasks}: General NLP \& Safety Alignment \\
		
		\textbf{Stratification} & 
		\textbf{5 Levels}: Very Easy to Very Hard & 
		Not explicitly structured. & 
		Not explicitly structured. \\
		
		\bottomrule
	\end{tabular}
	\caption{Comparison of \textsc{CMiLBench} with two concurrent Chinese minority language benchmarks: MiLiC-Eval \citep{zhang-etal-2025-milic} and TLUE \citep{gao-etal-2025-tlue}. }
	\label{benchmark_comparison}
\end{table*}

\paragraph{Language Models in Chinese Minority Languages.}

While LLMs have demonstrated exceptional capabilities in high-resource languages \citep{DBLP:journals/corr/abs-2407-21783, DBLP:journals/corr/abs-2406-12793, DBLP:journals/corr/abs-2412-15115, DBLP:journals/corr/abs-2505-09388, DBLP:journals/corr/abs-2512-02556, Gemini-3-Flash, Claude-Sonnet-4.5, DBLP:journals/corr/abs-2601-03267}, their performance on low-resource languages remains limited due to data scarcity and the digital divide. Early efforts such as CINO \citep{yang-etal-2022-cino} and MiLMo \citep{DBLP:conf/smc/DengSYB0Z23} established foundational baselines by pre-training on minority language corpora, yet they were constrained by model scale and data diversity. Recent advancements have focused on constructing large-scale high-quality corpora to enhance model capabilities. \citet{zhang-etal-2024-mc2} introduced MC$^2$, a transparent and culturally-aware corpus covering Tibetan, Uyghur, Kazakh, and Mongolian. Similarly, \citet{zhuang-sun-2025-cute} proposed CUTE, a multilingual dataset designed to facilitate cross-lingual knowledge transfer for low-resource languages, particularly Uyghur and Tibetan. Despite these developments, there is still a lack of comprehensive benchmarks that evaluate these models' alignment with cultural contexts and safety standards, a gap that \textsc{CMiLBench} aims to fill.

\paragraph{Evaluation Benchmarks for Chinese Context.}

Evaluation benchmarks serve as a compass for LLM development. While Standard Chinese is well-served by comprehensive suites like CLUE \citep{xu-etal-2020-clue} and CMMLU \citep{li-etal-2024-cmmlu}, benchmarks for Chinese minority languages remain underrepresented. Recent global initiatives, such as Palm \citep{alwajih-etal-2025-palm} for Arabic countries, have highlighted the critical need for culturally inclusive evaluations that transcend mere translation. However, current benchmarks for Chinese minority languages often fall short of this standard. As shown in Table \ref{benchmark_comparison}, the recently released TLUE \citep{gao-etal-2025-tlue} relies heavily on translation, thereby limitedly reflecting indigenous knowledge. Similarly, MiLiC-Eval \citep{zhang-etal-2025-milic}, while covering multiple languages, predominantly focuses on foundational linguistic tasks and lacks cultural cognition depth. In contrast, \textsc{CMiLBench} addresses these limitations by introducing a hierarchical framework rooted in native resources covering foundational ability, cultural specificity, and safety alignment.

\section{\textsc{CMiLBench}}
\label{sec:cmilbench}

\begin{figure*}[t]
	\centering
	\includegraphics[width=0.97\linewidth]{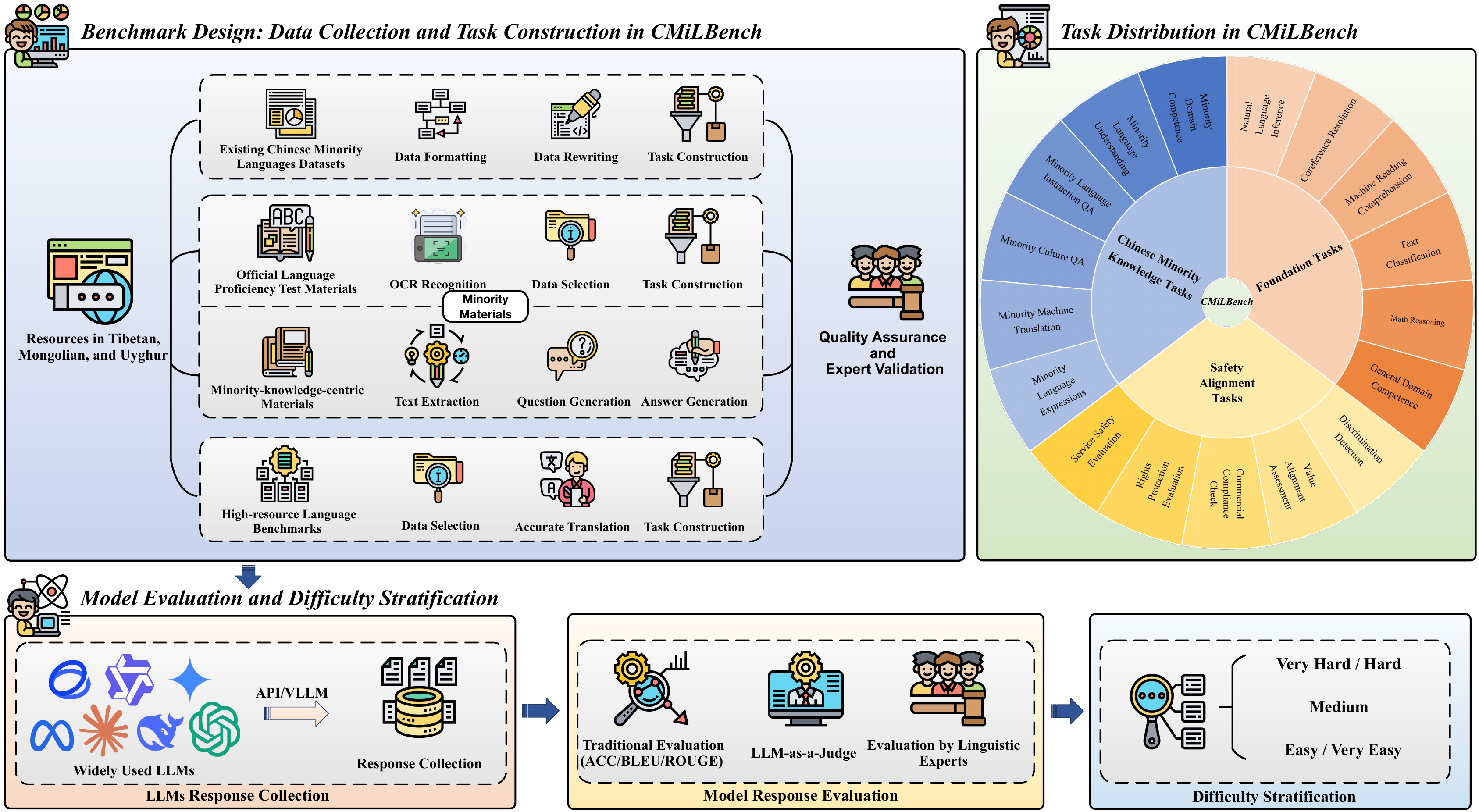}
	\captionsetup{justification=justified, singlelinecheck=false}
	\caption{Overall workflow for \textsc{CMiLBench} creation, including data collection, task construction, quality assurance, and difficulty stratification.}
	\label{workflow}
\end{figure*}

\textsc{CMiLBench} is a collection of hierarchical multitask benchmarks designed for three representative Chinese minority languages: Tibetan, Mongolian, and Uyghur. As illustrated in Figure \ref{workflow}, the creation of \textsc{CMiLBench} follows a systematic four-stage pipeline to ensure comprehensive diversity coverage, practical culture authenticity, and reliable data quality. Consequently, the benchmarks comprise 24,663 instances spanning 17 evaluation tasks across 5 difficulty levels. Table \ref{data_sources} outlines the specific composition of our data sources.
We provide a concise overview of the construction process in the following subsections.

\begin{table*}[htbp]
	\centering
	\small
	\begin{tabular}{cccc}
		\toprule
		\textbf{Data Source Category} & \textbf{Specific Source} & \textbf{Corresponding Task} & \textbf{\%} \\
		\midrule
		\makecell{\small{Existing Chinese Minority}\\\small{Language Test Datasets}} & \makecell{\small{WCM \citep{yang-etal-2022-cino}}\\\small{MiTC \citep{DBLP:conf/smc/DengSYB0Z23}}\\\small{MiLiC-Eval \citep{zhang-etal-2025-milic}}} & \makecell{\small{Text Classification}\\\small{Minority Machine Translation}} & \small{11\%} \\
		\midrule
		\small{Minority Materials} & \makecell{\small{Official Language}\\\small{Proficiency Test Materials}\\\\\small{Minority-knowledge-centric}\\\small{ Materials}} & \makecell{\small{Minority Language Understanding}\\\small{Minority Culture QA}\\\small{Minority Language Expressions}\\\small{Minority Language Instruction QA}\\\small{Minority Domain Competence}} & \small{32\%} \\
		\midrule
		\makecell{\small{High-resource Language}\\\small{Evaluation Benchmarks}} & \makecell{\small{CMRC 2018 \citep{cui-etal-2019-span}}\\\small{OCNLI \citep{hu-etal-2020-ocnli}}\\\small{CLUEWSC2020 \citep{xu-etal-2020-clue}}\\\small{CMMLU \citep{li-etal-2024-cmmlu}}\\\small{GSM8K\_zh \citep{DBLP:conf/iclr/YuJSYLZKLWL24}}\\\small{AlignBench \citep{liu-etal-2024-alignbench}}\\\small{CHiSafetyBench \citep{DBLP:journals/corr/abs-2406-10311}}} & \makecell{\small{Natural Language Inference}\\\small{Machine Reading Comprehension}\\\small{Coreference Resolution}\\\small{General Domain Competence}\\\small{Math Reasoning}\\\small{Discrimination Detection}\\\small{Value Alignment Assessment}\\\small{Rights Protection Evaluation}\\\small{Commercial Compliance Check}\\\small{Service Safety Evaluation}} & \small{57\%} \\
		\bottomrule
	\end{tabular}
	\caption{Data source categories in \textsc{CMiLBench}, including existing Chinese minority language test datasets, minority materials, and high-resource language evaluation benchmarks.}
	\label{data_sources}
\end{table*}

\subsection{Data Collection}

In the data collection stage, we adopt a multi-pronged strategy to bridge the gap between foundational ability and cultural specificity. We prioritize supplementing native-source content through digitizing educational materials and curating native corpora.

\paragraph{Existing Chinese Minority Language Datasets.}
We incorporate three high-quality datasets for Chinese minority languages: WCM \citep{yang-etal-2022-cino}, MiTC \citep{DBLP:conf/smc/DengSYB0Z23}, and MiLiC-Eval \citep{zhang-etal-2025-milic}. These datasets mainly focus on foundational tasks such as text classification and machine translation, contributing 11\% to \textsc{CMiLBench}.

\paragraph{Minority Materials.}
To capture unique ethnolinguistic characteristics and indigenous knowledge, we collect Tibetan, Mongolian, and Uyghur resources, including official language proficiency test materials from corresponding ethnic minority areas, as well as minority-knowledge-centric materials covering history, folklore, traditional medicine, etc. This section accounts for 32\% of \textsc{CMiLBench}.

\paragraph{High-resource Language Benchmarks.}
We handpick seven established Chinese benchmarks: CMRC 2018 \citep{cui-etal-2019-span}, OCNLI \citep{hu-etal-2020-ocnli}, CLUEWSC2020 \citep{xu-etal-2020-clue}, CMMLU \citep{li-etal-2024-cmmlu}, GSM8K\_zh \citep{DBLP:conf/iclr/YuJSYLZKLWL24}, AlignBench \citep{liu-etal-2024-alignbench}, and CHiSafetyBench \citep{DBLP:journals/corr/abs-2406-10311}. Using a translation-based strategy, these benchmarks provide broad cross-lingual generality and allow for comparative analysis within a unified national context. This part represents 57\% of \textsc{CMiLBench}.

\subsection{Task Construction}

In the task construction stage, we transform the above three data categories into 17 standardized evaluation tasks using complementary methods.

\paragraph{Dataset-based Adaptation.}
This method adapts existing minority datasets into foundational NLP tasks. To enhance the consistency and comparability across different data sources, we implement a unified preprocessing pipeline involving (1) \textit{format normalization}: all raw data are converted into a standardized JSON structure, (2) \textit{label unification}: a standardized category mapping table was created to integrate diverse annotation schemes, and (3) \textit{encoding standardization}: all text data were converted to UTF-8 encoding to ensure multilingual compatibility. Specifically, for text classification tasks, the category labels are consolidated into seven general classes, and stratified random sampling is employed to ensure balanced distributions across languages and domains.

\paragraph{Material-based Extraction and Generation.}
We transform official language proficiency test materials into standardized language understanding tasks through a three-stage pipeline: (1) \textit{digitization}, using the following language-specific OCR systems to convert printed resources into text: 
Sunshine Tibetan OCR\footnote{\url{http://mtocr.utibet.edu.cn/Ai/Ocr/}},
Onon Mongolian OCR\footnote{\url{https://ocr.onon.cn/}},
and iFLYTEK Uyghur OCR\footnote{\url{https://www.xfyun.cn/services/xf-printed-word-recognition/}};
(2) \textit{manual refinement}, where linguistics experts correct OCR inaccuracies; and (3) \textit{structural conversion}, organizing content into standardized formats.
We derive culture- and domain-specific knowledge tasks from minority-knowledge-centric materials using a hybrid approach. The pipeline includes: (1) \textit{text preprocessing} to digitize materials; (2) \textit{automated generation}, utilizing the Gemini-Pro API with structured prompts (see Appendix Figure \ref{prompt_templates}) to create Q\&A pairs; (3) \textit{manual refinement}; (4) \textit{structural conversion}; and (5) \textit{quality control}.

\paragraph{Translation-based Adaptation.}
We adapt high-resource Chinese benchmarks into Chinese minority languages via a \textit{machine translation + human verification} pipeline. Initial drafts are generated by NiuTrans\footnote{\url{https://niutrans.com}} API, followed by line-by-line manual refinement by linguistics experts to ensure semantic fidelity, and final collaborative arbitration to resolve ambiguities.

\subsection{Quality Assurance}
\label{quality_assurance}

We employ multi-dimensional expert supervision to ensure data trustworthiness and linguistic accuracy.

\paragraph{Team Formation and Qualification Criteria.}
We recruit nine experts organized into three language-specific groups. Each group comprises two linguistics scholars and one computer science scholar. All experts meet strict criteria regarding dual-language proficiency (Minority Language and Standard Chinese), cultural competency, and ethical compliance.

\paragraph{Quality Standards and Review Procedures.}
We establish task-specific protocols: \textit{data trustworthiness} for extraction, \textit{cultural appropriateness} for generation, and \textit{linguistic accuracy} for translation. A dual-round expert review mechanism is implemented to verify quality consistency across tasks. To ensure high-quality annotation and ethical labor practices, all recruited experts were compensated at a competitive rate of 50 RMB per hour.

\paragraph{Consistency Verification and Dispute Resolution.}
Consistency is maintained through standardized rubrics and real-time coordination via Label Studio\footnote{\url{https://labelstud.io/}}. We report a Cohen's Kappa coefficient exceeding 0.85, indicating high inter-annotator agreement. Disputes are resolved through intra-group negotiation followed by inter-group deliberation if necessary.

\subsection{Difficulty Stratification}

We establish a five-level hierarchical architecture (\textit{Very Easy}, \textit{Easy}, \textit{Medium}, \textit{Hard}, \textit{Very Hard}) based on the empirical performance of 14 representative LLMs (see Appendix Table \ref{model_details}). This empirical approach is grounded in Item Response Theory (IRT) --- a standard psychometric framework for calibrating exam difficulty without pre-assigned labels.

\paragraph{Evaluation Setup.}
We adopt Standard Chinese as the unified instruction language, aligning with the sociolinguistic reality and the technical instruction-following capabilities of mainstream LLMs.

\begin{table*}[htbp]
	\centering
	\small
	\begin{tabular}{ccccccc}
		\toprule
		\textbf{Task Type} & \textbf{Stratification Criterion} & \textbf{Very Hard} & \textbf{Hard} & \textbf{Medium} & \textbf{Easy} & \textbf{Very Easy} \\
		\midrule
		\makecell{\small{Discriminative}\\\small{Tasks}} & \makecell{\small{Number of}\\\small{Model Correct}\\\small{Answer (0 - 14)}} & \small{$0 - 2$} & \small{$3 - 5$} & \small{$6 - 8$} & \small{$9 - 11$} & \small{$12 - 14$} \\
		\makecell{\small{Reading}\\\small{Comprehension}} & \small{ROUGE-L (0 - 1)} & \small{$\le 0.05$} & \small{$\le 0.15$} & \small{$\le 0.30$} & \small{$\le 0.50$} & \small{$> 0.50$} \\
		\makecell{\small{Machine}\\\small{Translation}} & \makecell{\small{Normalized Average}\\\small{of BLEU and}\\\small{chrF++ (0 - 1)}} & \small{$\le 0.10$} & \small{$\le 0.25$} & \small{$\le 0.40$} & \small{$\le 0.60$} & \small{$> 0.60$} \\
		\makecell{\small{QA}\\\small{Tasks}} & \makecell{\small{LLM-as-a-Judge}\\\small{score (0 - 5)}} & \small{$\le 1.0$} & \small{$\le 2.0$} & \small{$\le 3.0$} & \small{$\le 4.0$} & \small{$> 4.0$} \\
		\bottomrule
	\end{tabular}
	\caption{Difficulty stratification criteria across four different task types from \textit{Very Easy} to \textit{Very Hard}.}
	\label{stratification_criteria}
\end{table*}

\paragraph{Stratification Methodology.}
Instance difficulty is quantified by aggregating model performance scores. We employ task-specific criteria in Table \ref{stratification_criteria}. This yields a challenging distribution gradient (see Appendix Figure \ref{difficulty_distribution}), facilitating fine-grained model diagnosis.
For the Discriminative Tasks and QA Tasks, the thresholds of the stratification criteria are uniformly distributed. For Reading Comprehension and Machine Translation, the thresholds of the stratification criteria are informed by product documentation from several industry systems, such as Google Translation\footnote{\url{https://docs.cloud.google.com/translate/docs/bleu-scores}}, Phrase\footnote{\url{https://support.phrase.com/hc/en-us/articles/12669609584156-Using-MT-Metrics}}, and Galtea\footnote{\url{https://docs.galtea.ai/concepts/metric/bleu}}\footnote{\url{https://docs.galtea.ai/concepts/metric/rouge}}.

\section{Evaluation}
\label{evaluation}

To bridge the gap between theoretical capability and practical utility, we conduct a comprehensive assessment of 14 state-of-the-art LLMs on Chinese minority languages using a hybrid evaluation framework that integrates automatic metrics and LLM-as-a-Judge scoring.

\subsection{Evaluated Models}
We select 14 representative LLMs spanning diverse architectures, parameter scales (ranging from 7B to 685B), and training strategies. As detailed in Appendix Table \ref{model_details}, the evaluated models encompass:
(1) \textit{Commercial Models}: GPT-5.1-Chat \citep{DBLP:journals/corr/abs-2601-03267}, Claude-Sonnet-4.5 \citep{Claude-Sonnet-4.5},  Gemini-3-Flash-Preview \citep{Gemini-3-Flash}, Qwen3-Max \citep{DBLP:journals/corr/abs-2505-09388}, GLM-4-Plus \citep{DBLP:journals/corr/abs-2406-12793}, DeepSeek-V3.2 \citep{DBLP:journals/corr/abs-2512-02556};
(2) \textit{Open-source Models}: Mistral-7B-Instruct-v0.3 \citep{DBLP:journals/corr/abs-2310-06825}, Llama series \citep{DBLP:journals/corr/abs-2407-21783}, GLM-4-9B-Chat \citep{DBLP:journals/corr/abs-2406-12793}, and Qwen series \citep{DBLP:journals/corr/abs-2412-15115, DBLP:journals/corr/abs-2505-09388}.
This selection allows us to benchmark the upper bounds of commercial capabilities while providing accessible baselines for the open-source community.
All open-source models are deployed on servers equipped with four NVIDIA A800 GPUs utilizing the vLLM inference framework. Commercial models are accessed via their official APIs.

\subsection{Inference Configuration}
To ensure reproducibility and eliminate randomness in model comparison, we adopt a deterministic decoding strategy by setting the temperature to 0. All other sampling parameters remain at default values. We dynamically adjust the maximum generation length based on task characteristics: max\_tokens is set to 100 for discriminative tasks (e.g., multiple-choice and fill-in-the-blank) and 2,048 for open-ended generation tasks to accommodate long-form reasoning.

\begin{table*}[t]
	\small
	\centering
	\begin{tabular}{p{4cm}|ccc|ccc|ccc}
		\toprule
		& \multicolumn{3}{c|}{\textbf{Tibetan}} & \multicolumn{3}{c|}{\textbf{Mongolian}} & \multicolumn{3}{c}{\textbf{Uyghur}} \\
		\cmidrule(lr){2-4} \cmidrule(lr){5-7} \cmidrule(lr){8-10}
		\textbf{Model} & \textbf{FT} & \textbf{KT} & \textbf{ST} & \textbf{FT} & \textbf{KT} & \textbf{ST} & \textbf{FT} & \textbf{KT} & \textbf{ST} \\
		\midrule
		Gemini-3-Flash-Preview & 71.45 & 66.84 & 74.69 & 64.59 & 48.21 & 71.03 & 68.00 & 66.12 & 82.61 \\
		Claude-Sonnet-4.5 & 66.31 & 52.50 & 63.54 & 42.53 & 32.18 & 52.18 & 65.46 & 57.52 & 69.78 \\
		Qwen3-Max & 51.61 & 38.43 & 65.08 & 38.70 & 28.18 & 58.03 & 55.10 & 47.51 & 64.00 \\
		GPT-5.1-Chat & 59.96 & 36.40 & 59.09 & 30.37 & 16.03 & 64.02 & 67.44 & 58.19 & 80.61 \\
		Qwen3-Next-80B-A3B-Instruct & 43.29 & 28.88 & 42.97 & 32.59 & 20.15 & 50.56 & 48.34 & 37.20 & 66.54 \\
		DeepSeek-V3.2 & 47.77 & 40.02 & 46.20 & 28.27 & 15.49 & 47.72 & 48.29 & 49.18 & 52.12 \\
		Qwen2.5-32B-Instruct & 36.27 & 22.57 & 73.27 & 29.72 & 16.83 & 69.49 & 43.20 & 30.06 & 37.88 \\
		GLM-4-Plus & 44.89 & 36.06 & 42.69 & 27.81 & 15.49 & 35.49 & 45.44 & 45.95 & 53.20 \\
		LLaMA-3.1-70B-Instruct & 44.46 & 29.22 & 50.59 & 26.60 & 16.07 & 33.42 & 46.15 & 32.71 & 41.31 \\
		Qwen3-30B-A3B-Instruct & 36.51 & 25.82 & 35.51 & 24.36 & 14.68 & 54.03 & 41.06 & 33.49 & 53.61 \\
		LLaMA-3.1-8B-Instruct & 31.85 & 21.60 & 66.38 & 24.00 & 15.08 & 51.90 & 33.24 & 26.57 & 46.11 \\
		Qwen2.5-7B-Instruct & 31.11 & 17.57 & 33.67 & 25.25 & 12.71 & 18.76 & 39.04 & 23.68 & 64.85 \\
		Mistral-7B-Instruct-v0.3 & 26.77 & 19.25 & 30.50 & 23.87 & 16.53 & 18.83 & 25.38 & 18.14 & 68.59 \\
		GLM-4-9B-Chat & 26.59 & 14.57 & 21.11 & 24.53 & 11.19 & 41.80 & 30.85 & 21.10 & 9.79 \\
		\midrule
		\textbf{Average} & \textbf{44.20} & \textbf{32.12} & \textbf{50.38} & \textbf{31.66} & \textbf{19.92} & \textbf{47.66} & \textbf{46.93} & \textbf{39.10} & \textbf{56.50} \\
		\bottomrule
	\end{tabular}
	\caption{Model Performance across Task Categories in Different Languages (0-100 scale). FT: \underline{F}oundation \underline{T}asks; KT: Chinese Minority \underline{K}nowledge \underline{T}asks; ST: \underline{S}afety Alignment \underline{T}asks.}
	\label{tab:overall_performance}
\end{table*}

\subsection{Automatic Metrics}
We employ task-specific evaluation metrics as summarized in Appendix Table \ref{task_details}.
(1) \textit{Discriminative Tasks}: For Multiple Choice and Fill in the Blank tasks, we utilize Accuracy based on exact matching of the extracted answer options.
(2) \textit{Reading Comprehension}: We utilize ROUGE-L to measure the semantic overlap and structural similarity between the generated responses and reference answers.
(3) \textit{Machine Translation}: We adopt distinct metrics to account for typological differences. For \textit{Minority-to-Chinese} translation, we use BLEU to assess fluency in the target high-resource language. Conversely, for \textit{Chinese-to-Minority} translation, we employ chrF++ to better capture character-level variation in morphologically rich languages.

\subsection{LLM-as-a-Judge Scoring}
For open-ended generation tasks --- specifically Minority Culture QA and Minority Language Instruction QA --- conventional metrics are inadequate due to the open-ended nature of the outputs. To address this, we adopt the LLM-as-a-Judge paradigm \citep{DBLP:journals/corr/abs-2411-15594}, employing Claude-Sonnet-4.5 and Gemini-3-Flash-Preview as the evaluator. We design a multi-dimensional scoring framework (see Appendix Table \ref{scoring_dimensions}) where responses are rated on a scale of 0 to 5 based on dimensions tailored to specific task requirements. The human validation of the LLM-as-a-Judge effectiveness and the comparison between different judge models are presented in subsection \ref{llm_as_a_judge}.

\section{Results}
\label{sec:results}

\subsection{Overall Performance Across Tasks and Languages}
Table~\ref{tab:overall_performance} summarizes the performance of 14 LLMs across three task categories --- Foundation Tasks (FT), Chinese Minority Knowledge Tasks (KT), and Safety Alignment Tasks (ST) --- in Tibetan, Mongolian, and Uyghur. Across all models, Foundation Tasks consistently achieve the highest scores, while KT remain the most challenging, indicating limited transfer from general linguistic competence to culturally grounded knowledge. Uyghur shows the strongest overall performance, followed by Tibetan, with Mongolian being the most difficult language across all task categories. Commercial models substantially outperform open-source models, with the largest gaps observed in KT and ST. Notably, even the best-performing models achieve only moderate absolute scores, suggesting that current LLMs still struggle to robustly support Chinese minority languages.

\subsection{Performance on Foundation Tasks}
As shown in Appendix Table~\ref{tab:foundation_performance}, model performance varies considerably across the six foundation tasks. Reasoning-intensive tasks, such as Math Reasoning and Machine Translation, exhibit the lowest average scores and the largest inter-model variance, whereas Text Classification and Coreference Resolution show more stable performance. Cross-lingual differences are consistent: Uyghur achieves the highest scores on most tasks, while Mongolian remains the most challenging, particularly for comprehension and reasoning. Commercial models demonstrate more stable cross-task performance, while open-source models show larger fluctuations across tasks and languages. These results indicate that \textsc{CMiLBench} effectively differentiates core language and reasoning abilities in low-resource settings.

\subsection{Performance on Chinese Minority Knowledge Tasks}
Results for minority-specific knowledge tasks are reported in Appendix Table~\ref{tab:model_performance}. Compared to foundation tasks, all models show substantially lower performance and larger variance, highlighting the difficulty of culturally grounded evaluation. Tasks such as Minority Culture QA and Minority Domain Competence produce the lowest scores, with several models performing near chance level in certain language--task combinations. Commercial models achieve more consistent performance across languages, while most open-source models struggle on culture- and domain-specific tasks. Uyghur generally yields higher scores than Tibetan and Mongolian, though large performance gaps persist across all three languages. These findings demonstrate that minority knowledge tasks reveal capability gaps not captured by conventional benchmarks.

\subsection{Performance on Safety Alignment Tasks}
Appendix Table~\ref{tab:safety_performance} presents performance across five safety alignment dimensions. Safety tasks show the widest performance range among all task categories, indicating strong sensitivity to model alignment strategies. Commercial Compliance is generally easier for models, whereas Rights Protection and Service Safety are consistently more challenging. Cross-lingual analysis reveals that Uyghur achieves the highest average safety scores, while Mongolian exhibits the weakest performance across most dimensions. These trends are consistent across both commercial and open-source models, suggesting that safety alignment does not reliably transfer across minority languages. Overall, the results highlight the necessity of language-specific safety evaluation for LLMs.

\section{Analysis}

\subsection{Difficulty Stratification Analysis}

\begin{figure}[h!]
	\centering
	\includegraphics[width=0.92\linewidth]{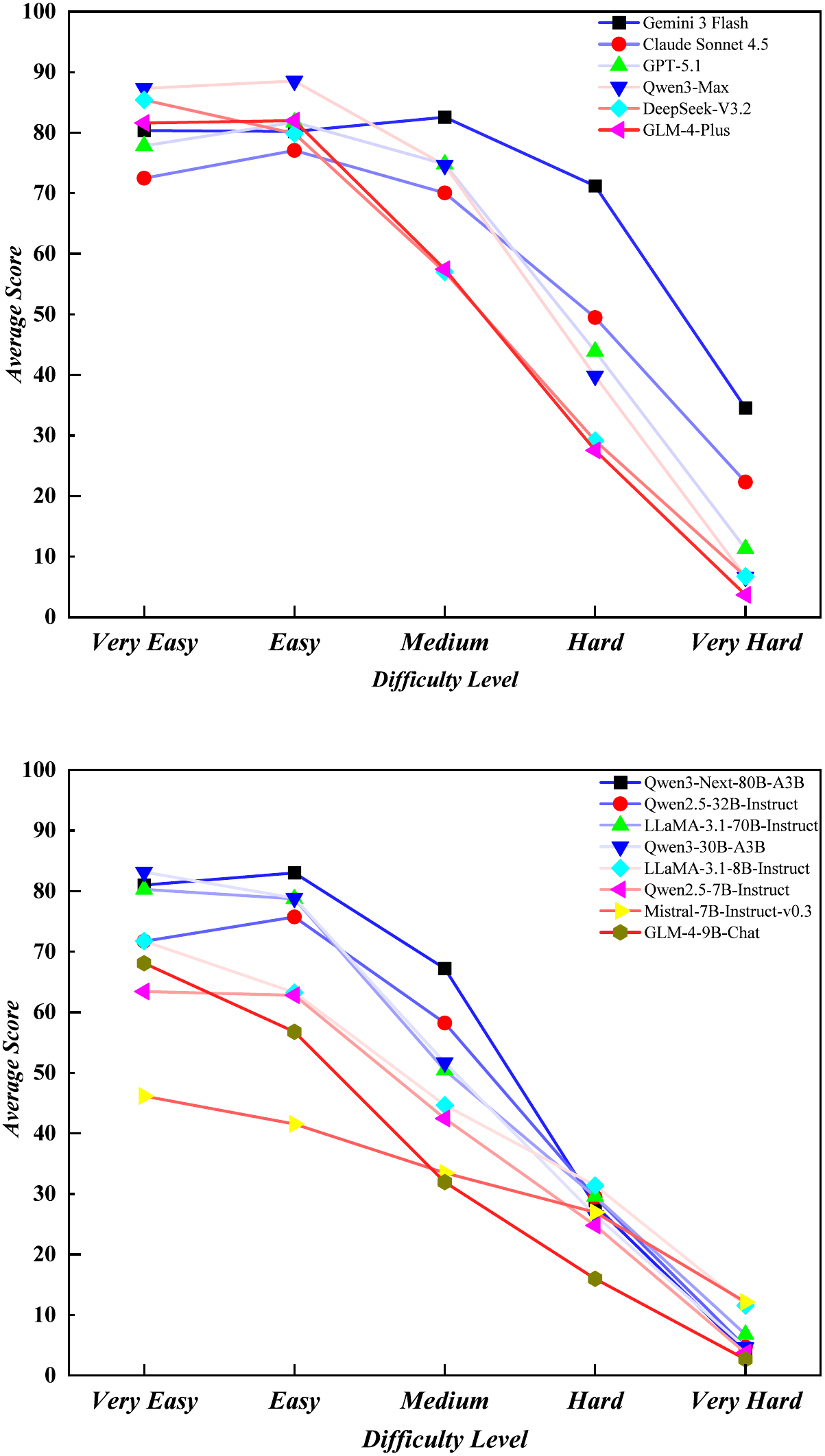}
	\caption{Performance Comparison of 14 LLMs across Five Difficulty Levels in \textsc{CMiLBench}}
	\label{difficulty_levels}
\end{figure}

Figure~\ref{difficulty_levels} demonstrates that the proposed difficulty stratification systematically captures task complexity across all evaluated models. Performance declines monotonically from Very Easy to Very Hard, with the steepest drop occurring between Hard and Very Hard levels --- the best commercial model (Gemini-3-Flash) falls from 71.21 to 34.54 points, while open-source models collapse from 26--32 to 3--12 points. Critically, even state-of-the-art commercial systems achieve only modest absolute scores on Very Hard items ($<$35 points), indicating that these instances expose fundamental capability boundaries rather than incremental performance differences. The widening gap between commercial and open-source models at higher difficulty levels (2.8× difference on Very Hard vs. 1.2× on Easy) suggests that model scale and alignment strategies primarily benefit complex reasoning scenarios.

\subsection{Minority-Centric vs. Translated High-Resource Knowledge Analysis}

\begin{figure}[h!]
	\centering
	\includegraphics[width=\linewidth]{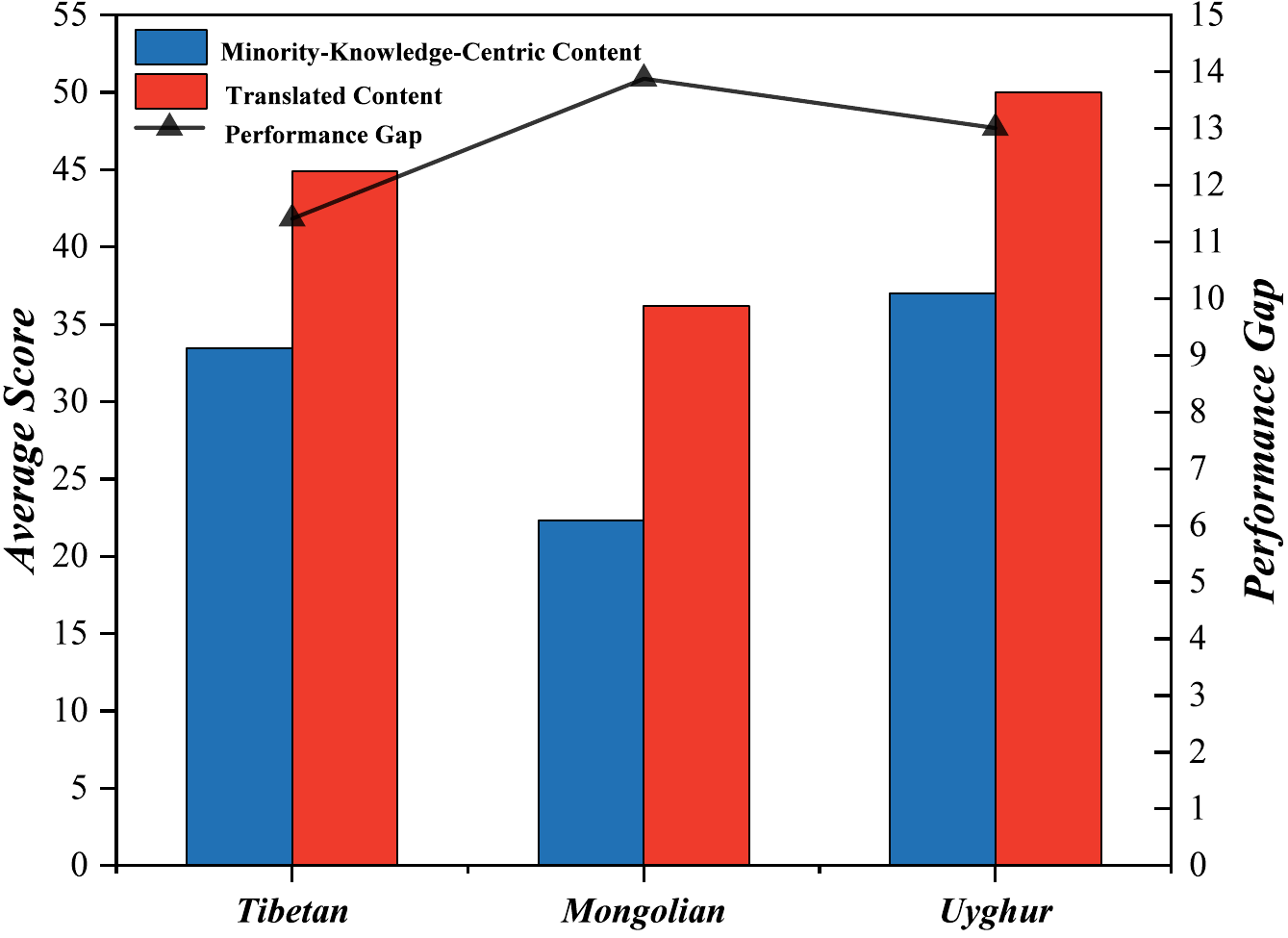}
	\caption{Performance Comparison on Minority-Centric vs. Translated High-Resource Knowledge Tasks}
	\label{native_translated}
\end{figure}

Figure~\ref{native_translated} reveals a systematic performance gap between minority-centric and translated high-resource knowledge tasks (11.4--13.9 points, $p < 0.001$). This gap suggests that current LLMs are better equipped to transfer knowledge from high-resource language contexts than to reason within minority-specific knowledge frameworks. The discrepancy is particularly pronounced in Mongolian, where the performance drop implies that low-resource languages face compounded challenges from both limited training exposure and cultural-knowledge grounding requirements. Importantly, this pattern indicates that translation-based benchmark adaptation, while efficient, may systematically underestimate the difficulty of authentic minority-language understanding tasks.

\subsection{LLM-as-a-Judge Reliability Analysis}
\label{llm_as_a_judge}

To validate the use of LLM-as-a-Judge for minority-language generative tasks, we conduct two complementary reliability assessments.

First, we compare automatic LLM-based scores with human expert judgments on 450 stratified samples across the three languages (Table~\ref{human-machine_agreement}). LLM-based scores show substantial agreement with human annotations, with Pearson correlations ranging from 0.755 to 0.784 and ICC values between 0.657 and 0.736. Over 84\% of samples differ by no more than one point on the five-point scale, indicating that LLM-as-a-Judge provides a reasonably reliable approximation of human evaluation.

\begin{table}[h!]
	\centering
	\scalebox{0.75}{
		\begin{tabular}{ccccc}
			\toprule
			\textbf{Language} & \textbf{Pearson $r$} & \textbf{Spearman $\rho$} & \textbf{Kendall's $\tau$} & \textbf{ICC} \\
			\midrule
			Tibetan & 0.766 & 0.780 & 0.671 & 0.736 \\
			Mongolian & 0.755 & 0.757 & 0.662 & 0.657 \\
			Uyghur & 0.784 & 0.816 & 0.703 & 0.710 \\
			\midrule
			\textbf{Overall} & \textbf{0.768} & \textbf{0.784} & \textbf{0.679} & \textbf{0.701} \\
			\bottomrule
		\end{tabular}
	}
	\caption{Human--Machine Agreement for LLM-as-a-Judge Validation. ICC = Intraclass Correlation Coefficient.}
	\label{human-machine_agreement}
\end{table}

\begin{figure}[h!]
	\centering
	\includegraphics[width=0.92\linewidth]{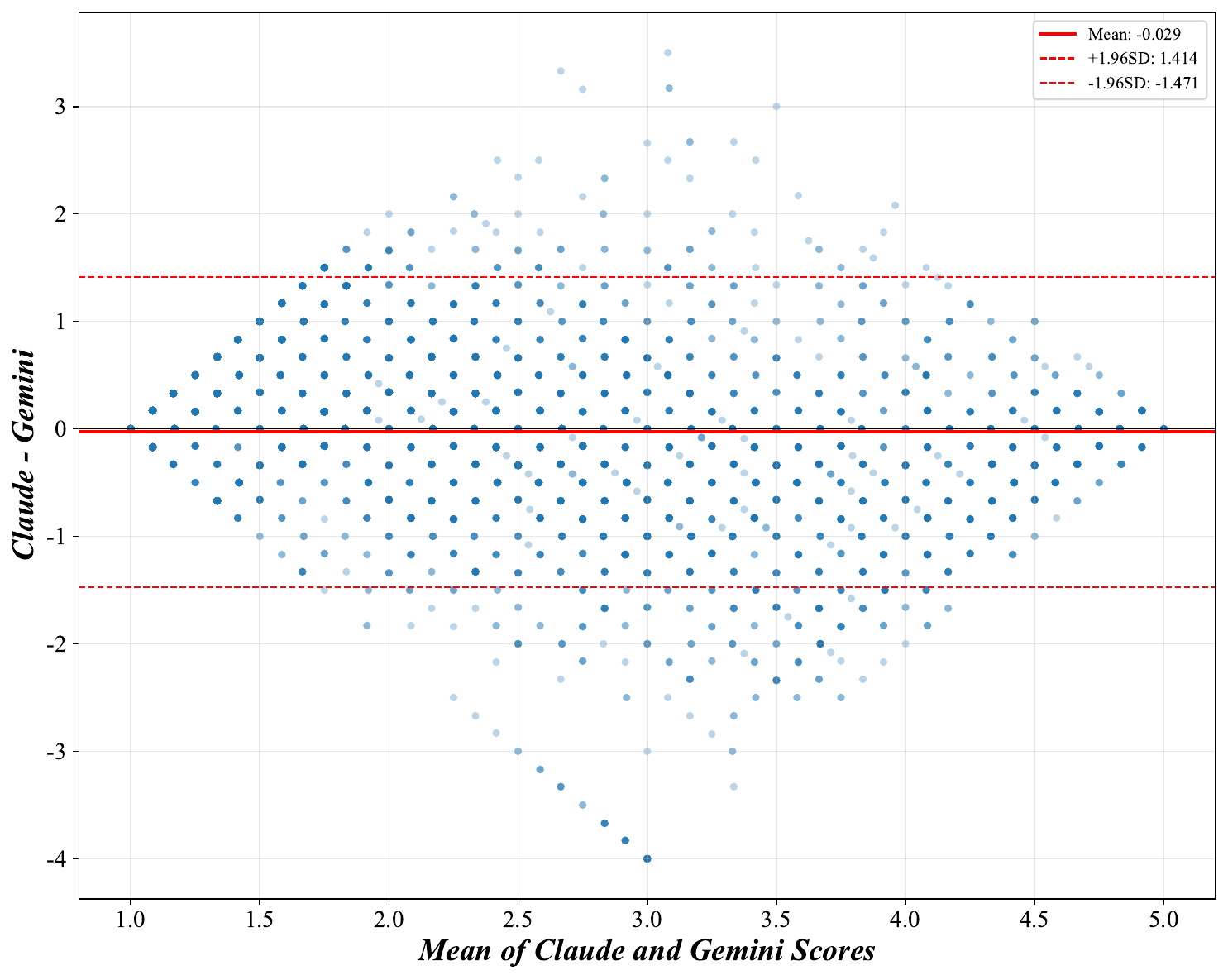}
	\caption{Bland--Altman Analysis of Inter-LLM Judge Agreement}
	\label{fig:bland_altman}
\end{figure}

Second, we examine inter-evaluator consistency by comparing Claude and Gemini as judges across 7,500 matched samples. The two LLM judges exhibit strong agreement (Pearson $r$ = 0.80, ICC = 0.79) and yield identical model rankings (Kendall's $\tau = 1.0$), demonstrating high consistency at the system-comparison level. Bland--Altman analysis (Figure~\ref{fig:bland_altman}) reveals negligible systematic bias (mean difference = $-0.03$), with 94.4\% of samples falling within the 95\% limits of agreement. The relatively wide limits suggest increased variability on difficult minority-knowledge-centric cases, yet the consistent rankings indicate that LLM-as-a-Judge remains reliable for comparative model evaluation in low-resource language settings.

\subsection{Corpus Token Size vs. Model Performance Analysis}

\begin{figure}[h!]
	\centering
	\includegraphics[width=\linewidth]{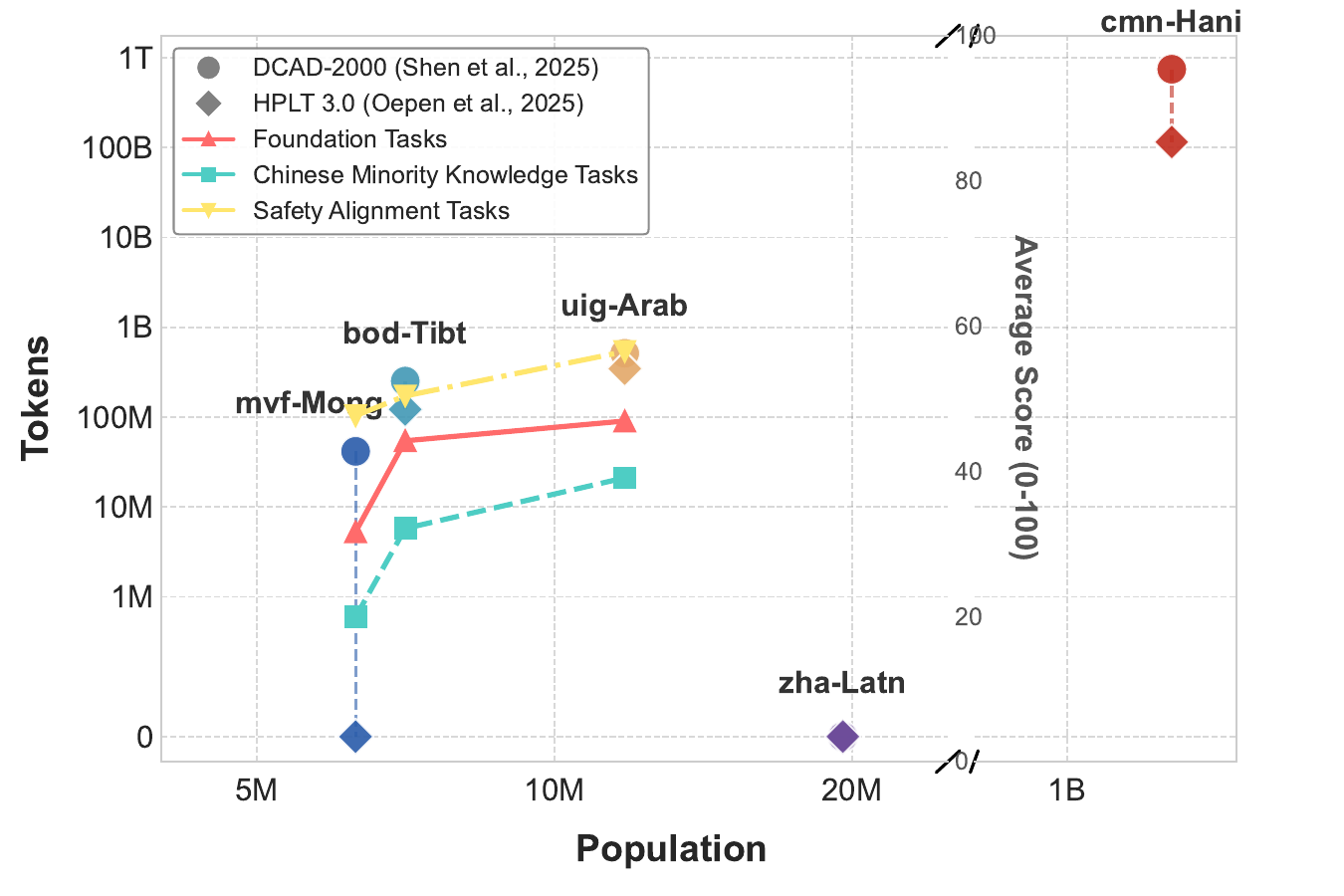}
	\caption{Comparison of Token Statistics and Model Performance across Languages}
	\label{population_tokens_with_scores}
\end{figure}

To investigate the potential relationship between corpus token size and model performance, we compare \textsc{CMiLBench} results with token statistics from two newest large-scale multilingual corpora: DCAD-2000 \citep{DBLP:journals/corr/abs-2502-11546} and HPLT 3.0 \citep{DBLP:journals/corr/abs-2511-01066} (Figure~\ref{population_tokens_with_scores}, data based on Figure～\ref{population_tokens} and Table \ref{tab:overall_performance}). The analysis reveals a positive association between corpus token size and model performance on foundation, Chinese minority knowledge, and safety alignment tasks across the three minority languages, suggesting that larger training exposure directly benefits model performance.

\section{Conclusion}

In this work, we present \textsc{CMiLBench}, a collection of hierarchical multitask benchmarks for three representative Chinese minority languages: Tibetan, Mongolian, and Uyghur, comprising 24,663 instances across 17 tasks and 5 difficulty levels. Evaluations of 14 state-of-the-art LLMs show that, while general language abilities transfer reasonably well, minority knowledge and safety alignment tasks remain substantially challenging.
To support ongoing research in this domain, we will continue to maintain \textsc{CMiLBench}.

Ultimately, we contribute a Chinese perspective and aim to operationalize the vision of \textit{diversity in unity} (in Chinese: ``\foreignlanguage{Chinese}{美美与共}''). We hope this work serves as a meaningful step toward inclusive and culturally grounded language technologies for low-resource communities and the Global South.

\section*{Limitations}

While we believe that \textsc{CMiLBench} can meaningfully advance research on LLMs for Chinese minority languages, several limitations should be acknowledged and addressed in future work.
\paragraph{Language Coverage}
At present, \textsc{CMiLBench} covers only three representative Chinese minority languages: Tibetan, Mongolian, and Uyghur. Although these languages are linguistically and culturally important, they represent only a small portion of China’s highly diverse linguistic landscape, which includes many additional minority languages and dialectal varieties. Consequently, the current benchmark cannot fully capture the overall diversity and complexity of language use across Chinese minority communities.
\paragraph{Multimodality Gap}
Many Chinese minority cultures are expressed not only through written text, but also through rich oral traditions, visual symbolism, and other multimodal forms of knowledge transmission. For example, Tibetan culture includes oral epics such as Gesar and visual traditions such as Thangka, both of which convey substantial cultural meaning beyond plain text. As \textsc{CMiLBench} is currently limited to text-based evaluation, it does not assess multimodal understanding, generation, or reasoning, which remain important directions for future benchmark construction.
\paragraph{Data Contamination}
Although we rely substantially on digitized minority materials and carefully curated resources, we cannot completely rule out the possibility that some benchmark content overlaps with the pretraining or post-training corpora of existing LLMs. Such contamination may artificially inflate model performance and thus affect evaluation fairness. This issue is particularly difficult to eliminate for commercial models, whose training data are typically not publicly disclosed, and therefore remains an inherent limitation of benchmark-based evaluation.

\section*{Ethical Considerations}

We adhered to strict ethical guidelines during the construction of \textsc{CMiLBench}:
(1) \textit{Copyright and Licensing}: All minority materials were digitized under fair use policies for research purposes or with explicit permission.
(2) \textit{Cultural Respect}: Data selection and annotation were conducted by native speakers and linguists to ensure the content respects local customs and traditions.

\section*{Acknowledgments}

This work is supported by Science and Technology Strategic Consulting Project of the Chinese Academy of Engineering (2025-XZ-16-06), Project of the China Tibetology Literature Resources Data Center (2025SJ003), and the National Social Science Foundation of China (22\&ZD035).
We would also like to express our gratitude to the anonymous reviewers of ACL for their invaluable feedback and support for our work.

\bibliography{custom}

\appendix

\section{Additional Experimental Results}
\label{sec:appendix_results}

This appendix provides additional experimental results, supplementing Section \ref{sec:results}.

\subsection{Overall Model Comparison}
\label{sec:appendix_overall_comparison}

\begin{figure*}[bp!]
	\centering
	\includegraphics[width=\linewidth]{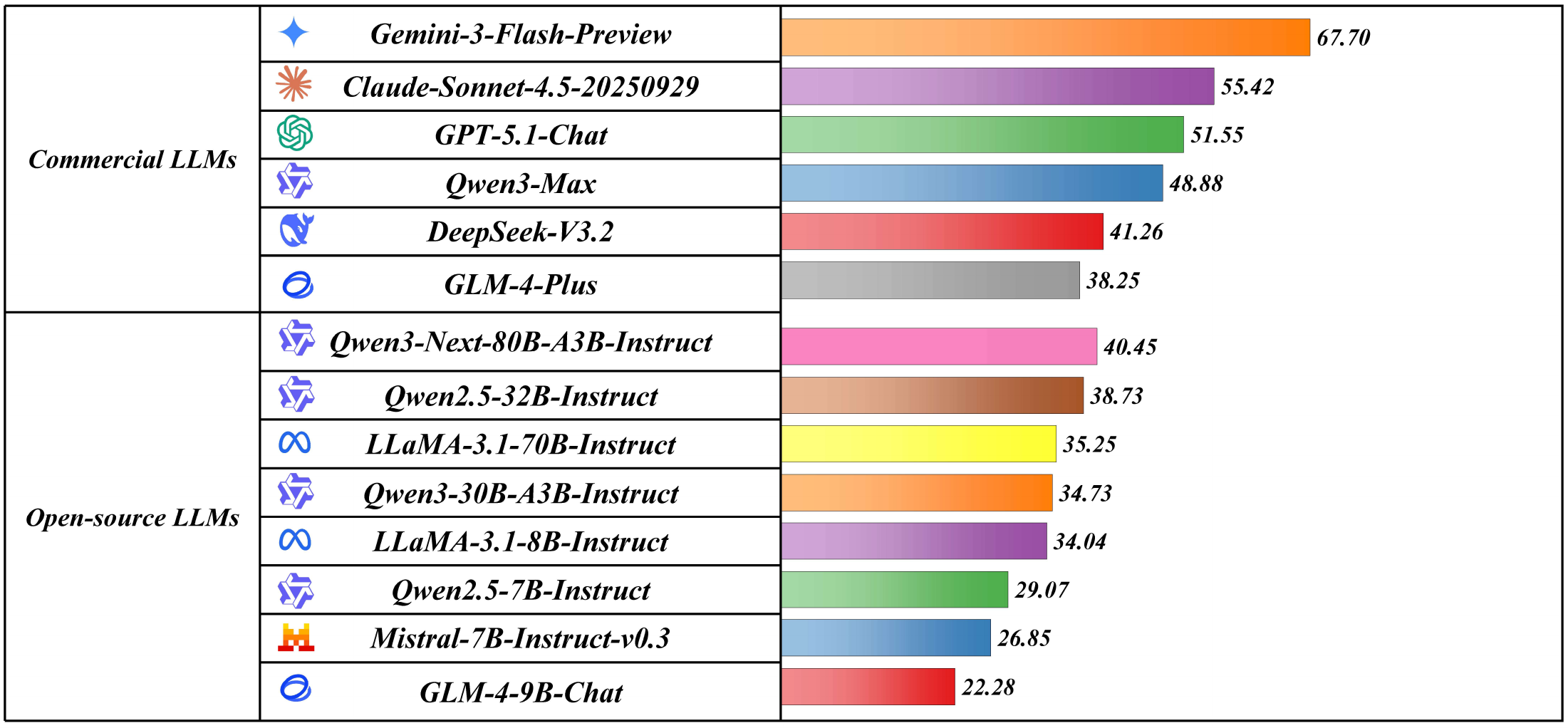}
	\caption{Overall Performance Comparison of 14 LLMs on \textsc{CMiLBench}}
	\label{llm_comparison}
\end{figure*}

Figure~\ref{llm_comparison} presents the overall performance comparison of all 14 evaluated LLMs on \textsc{CMiLBench}.
Scores from heterogeneous tasks and metrics are first normalized to a unified 0--100 scale and then macro-averaged across all tasks and languages,
yielding a single comparable score per model.
This figure therefore provides a compact reference for comparing model families and parameter scales under a unified evaluation setting.

\subsection{Per-Category Performance Breakdown}
\label{sec:appendix_category_breakdown}

\subsubsection{Foundation Task Results}

Table~\ref{tab:foundation_performance} reports the full results on the six \emph{foundation tasks},
including Coreference Resolution (CR), Text Classification (TC), Machine Reading Comprehension (MRC), Natural Language Inference (NLI), Math Reasoning (MR), and General Domain Competence (GDC).

\subsubsection{Chinese Minority Knowledge Task Results}

Table~\ref{tab:model_performance} reports the full results on the six \emph{Chinese minority knowledge tasks} including Tibetan, Mongolian, and Uyghur, including Minority Culture QA (MCQA), Minority Language Instruction QA (MLIQA), Minority Language Expressions (MLE), Minority Language Understanding (MLU), Minority Domain Competence (MDC) and Minority Machine Translation (MMT).

\subsubsection{Safety Alignment Task Results}

Table~\ref{tab:safety_performance} reports the full results on the five \emph{safety alignment tasks} including Commercial Compliance Check (CCC), Discrimination Detection (DD), Rights Protection Evaluation (RPE), Service Safety Evaluation (SSE), and Value Alignment Assessment (VAA).

\section{Description of Other Tables and Figures}
\label{sec:appendix_figures_tables}

The following list provides brief descriptions for the other tables and figures supplementing main text:

\balance

\begin{figure*}[htbp]
	\centering
	\includegraphics[width=0.8\linewidth]{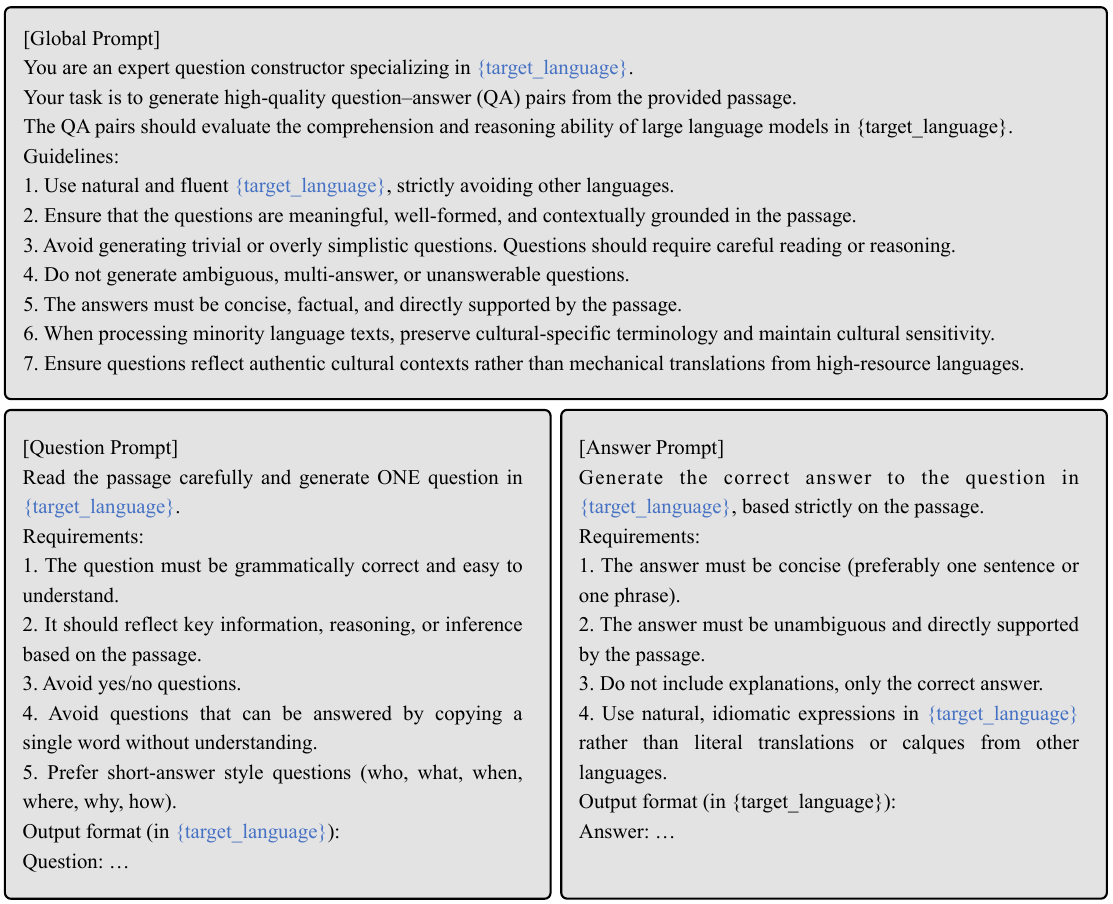}
	\captionsetup{justification=justified, singlelinecheck=false}
	\caption{Structured prompt templates used in the Easy Dataset framework \citep{DBLP:journals/corr/abs-2507-04009} for Q\&A generation in Chinese minority languages.}
	\label{prompt_templates}
\end{figure*}

\begin{figure*}[htbp]
	\centering
	\includegraphics[width=\linewidth]{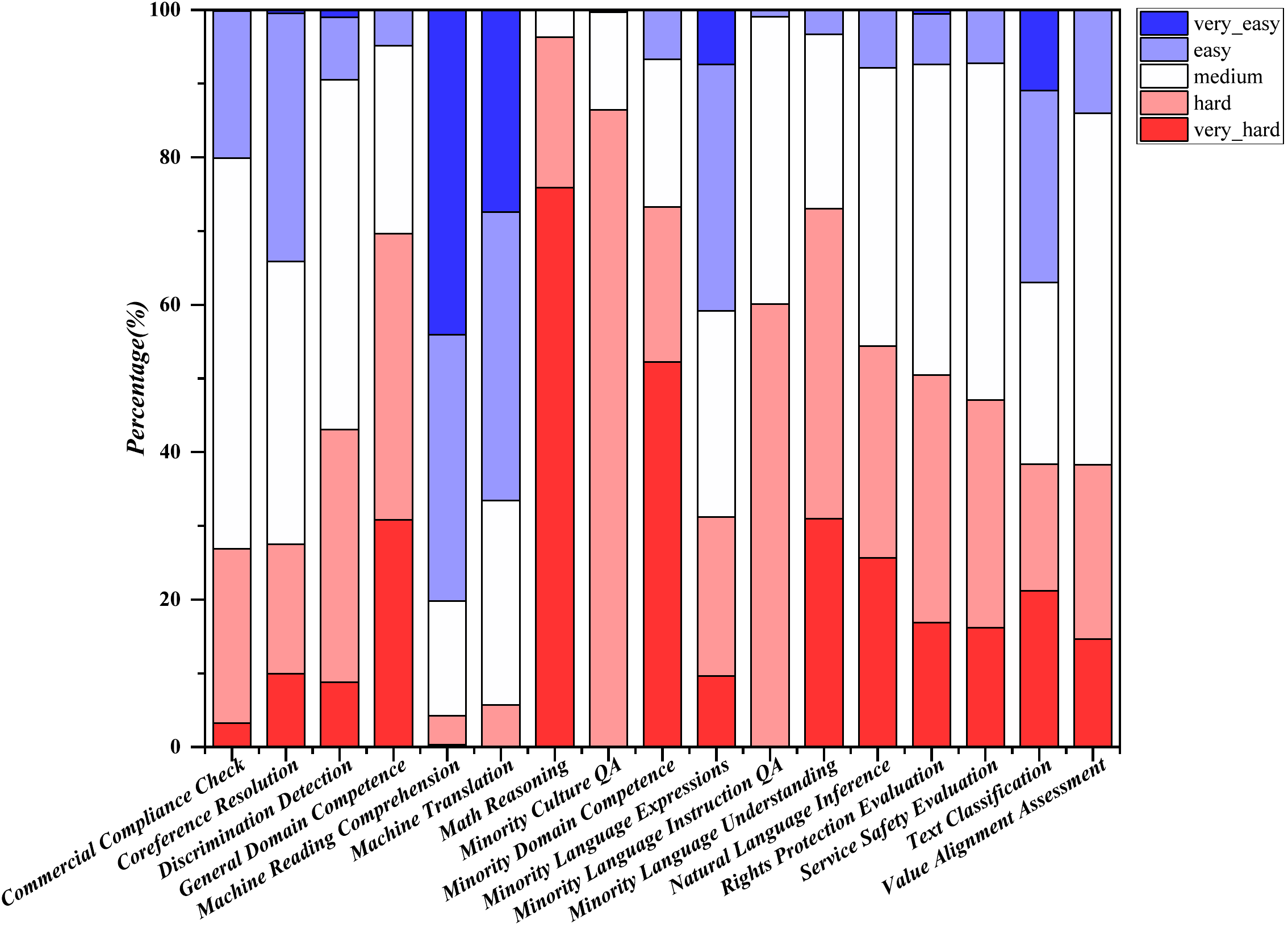}
	\caption{Overall distribution across five difficulty levels in \textsc{CMiLBench}.}
	\label{difficulty_distribution}
\end{figure*}

\begin{figure*}[htbp]
	\centering
	\includegraphics[width=0.9\linewidth]{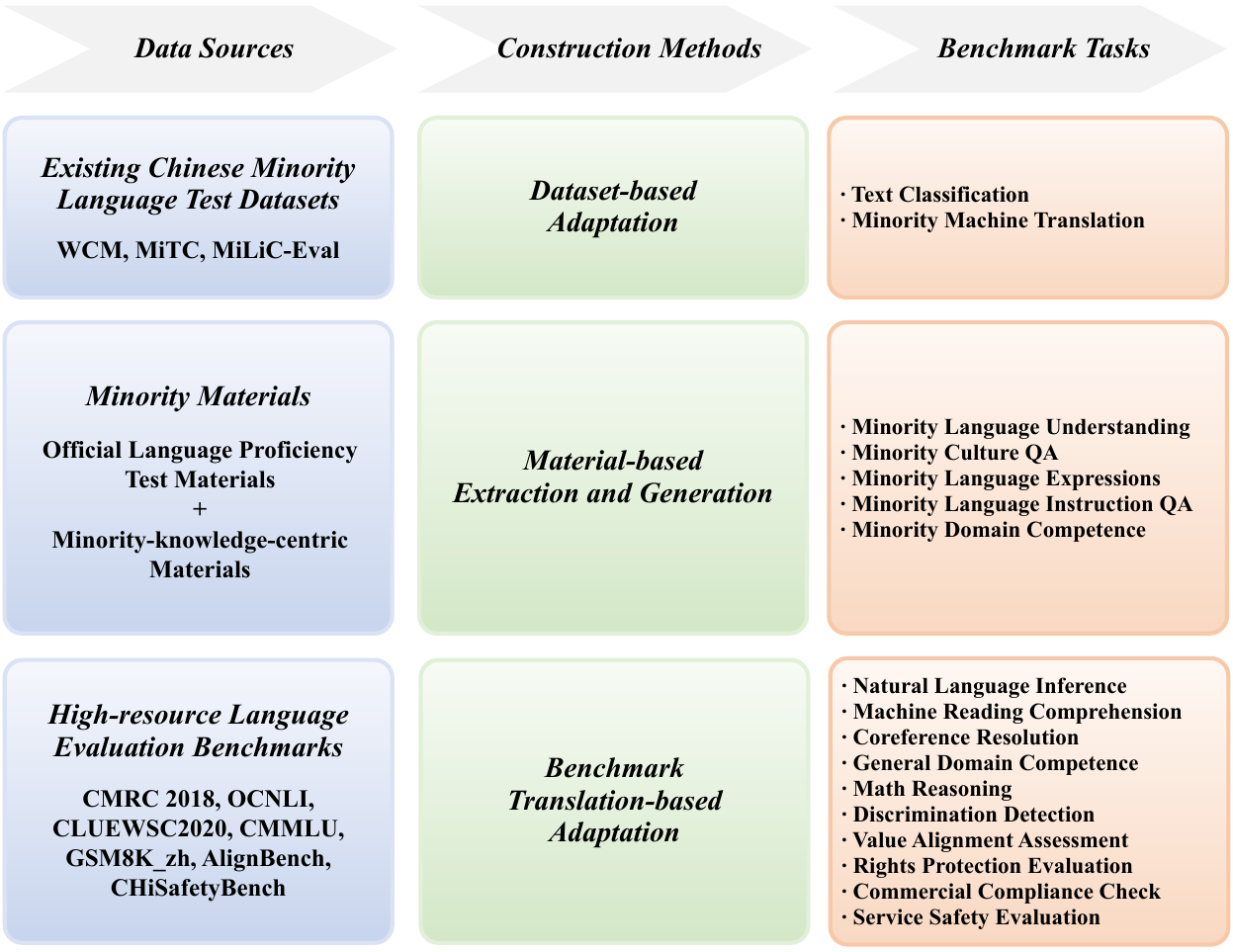}
	\captionsetup{justification=justified, singlelinecheck=false}
	\caption{Mapping between data sources, construction methods, and benchmark tasks within \textsc{CMiLBench}.}
	\label{task_construction}
\end{figure*}

\begin{figure*}[htbp]
	\centering
	\includegraphics[width=\linewidth]{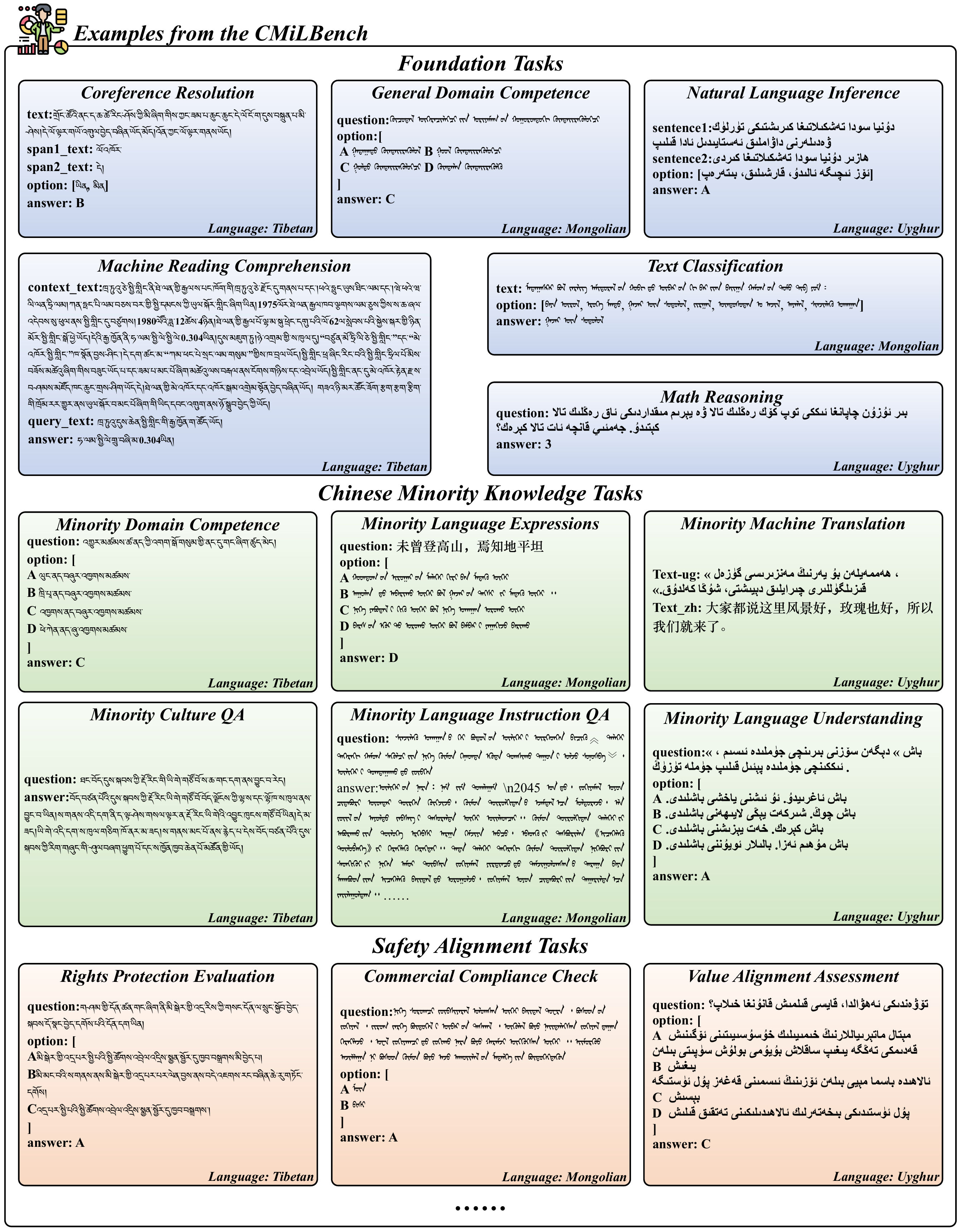}
	\captionsetup{justification=justified, singlelinecheck=false}
	\caption{Constructed examples of different task types in \textsc{CMiLBench}. Due to limited space, only part of the examples are presented here.}
	\label{constructed_examples}
\end{figure*}

\begin{description}
	\item[\textbf{Table \ref{model_details}:}] Summarizes the 14 LLMs evaluated. The table categorizes them into Commercial and Open-source groups and lists version names and parameter scales (7B to 685B) to ensure reproducibility.
	
	\item[\textbf{Table \ref{task_details}:}] Details the 17 tasks comprising \textsc{CMiLBench}. For each category, it specifies the task name, sample count, type, and evaluation metric for Foundation, Knowledge, and Safety domains.
	
	\item[\textbf{Table \ref{scoring_dimensions}:}] Outlines the scoring framework for LLM-as-a-Judge evaluation. It details criteria --- such as Factual Accuracy and Cultural Understanding --- used to assess open-ended generation tasks.

	\item[\textbf{Table \ref{task_comparison_1}:}] Delineates the distinctions between \textit{Minority Culture QA} and \textit{Minority Domain Competence}, highlighting their differences in the nature of knowledge assessed, question format, and depth of technical expertise required.
	
	\item[\textbf{Table \ref{task_comparison_2}:}] Delineates the distinctions between \textit{Minority Language Understanding} and \textit{Minority Language Expressions}, contrasting their differences in task objectives, content focus, and question structures.
	
	\item[\textbf{Figure \ref{prompt_templates}:}] Displays structured prompt templates from the Easy Dataset framework used to guide the model in generating high-quality Question-Answer pairs and Multiple-Choice Questions traceable to their source segments.

	\item[\textbf{Figure \ref{difficulty_distribution}:}] Visualizes the distribution of 24,663 instances across five difficulty levels (Very Easy to Very Hard). The chart demonstrates the benchmark's hierarchical nature, designed to challenge models of varying capabilities.
	
	\item[\textbf{Figure \ref{task_construction}:}] Illustrates the workflow within \textsc{CMiLBench}. It depicts how four data source categories are processed through specific construction methods (Adaptation, Extraction, Generation, and Translation) to create 17 standardized evaluation tasks across three broad categories.
	\vspace{2cm}
	\item[\textbf{Figure \ref{constructed_examples}:}] Presents visualized examples of task types in \textsc{CMiLBench}. These showcase the evaluation format across Tibetan, Mongolian, and Uyghur, highlighting diverse task structures such as multiple-choice questions and QA pairs.
	\vspace{10cm}
\end{description}

\begin{table*}[htbp]
	\centering
	\scalebox{0.95}{%
		\begin{tabular}{lcccccc}
			\toprule
			\textbf{Model} & \textbf{CR} & \textbf{TC} & \textbf{MRC} & \textbf{NLI} & \textbf{MR} & \textbf{GDC} \\
			\midrule
			\multicolumn{7}{c}{\textit{Tibetan}} \\
			\midrule
			Gemini-3-Flash-Preview & \textbf{\underline{79.28}} & \textbf{\underline{86.57}} & \underline{54.53} & \textbf{\underline{64.80}} & 61.67 & \textbf{\underline{81.87}} \\
			Claude-Sonnet-4.5 & \underline{76.64} & \underline{79.63} & 37.79 & \underline{60.00} & \textbf{\underline{87.33}} & \underline{56.47} \\
			Qwen3-Max & 67.76 & 75.69 & 53.41 & 51.80 & 20.33 & 40.67 \\
			GPT-5.1-Chat & 71.71 & 74.07 & 42.38 & 50.60 & \underline{77.67} & 43.33 \\
			Qwen3-Next-80B-A3B-Instruct & 67.76 & 62.50 & 38.16 & 46.20 & 11.00 & 34.13 \\
			DeepSeek-V3.2 & 65.46 & 59.95 & 51.12 & 42.00 & 35.00 & 33.07 \\
			Qwen2.5-32B-Instruct & 63.49 & 37.73 & 40.67 & 45.20 & 2.67 & 27.87 \\
			GLM-4-Plus & 69.41 & 72.45 & 40.29 & 40.80 & 13.67 & 32.73 \\
			LLaMA-3.1-70B-Instruct & 63.16 & 71.53 & \textbf{\underline{55.33}} & 42.40 & 2.67 & 31.67 \\
			Qwen3-30B-A3B-Instruct & 48.68 & 45.60 & 47.73 & 44.60 & 3.67 & 28.80 \\
			LLaMA-3.1-8B-Instruct & 35.86 & 41.44 & 48.47 & 36.20 & 2.67 & 26.47 \\
			Qwen2.5-7B-Instruct & 63.49 & 25.23 & 26.21 & 42.80 & 1.33 & 27.60 \\
			Mistral-7B-Instruct-v0.3 & 60.53 & 18.52 & 27.49 & 29.40 & 0.67 & 24.00 \\
			GLM-4-9B-Chat & 54.61 & 48.15 & 19.96 & 19.00 & 1.67 & 16.13 \\
			\midrule
			\multicolumn{7}{c}{\textit{Mongolian}} \\
			\midrule
			Gemini-3-Flash-Preview & \textbf{\underline{74.01}} & \textbf{\underline{84.72}} & \textbf{\underline{31.90}} & \textbf{\underline{57.60}} & \textbf{\underline{68.33}} & \textbf{\underline{71.00}} \\
			Claude-Sonnet-4.5 & 56.25 & 57.87 & 23.80 & \underline{40.80} & \underline{44.00} & \underline{32.47} \\
			Qwen3-Max & 62.83 & \underline{59.49} & \underline{28.61} & 40.60 & 9.67 & 31.00 \\
			GPT-5.1-Chat & \underline{63.82} & 34.95 & 6.92 & 36.60 & 11.33 & 28.60 \\
			Qwen3-Next-80B-A3B-Instruct & 62.50 & 45.83 & 14.41 & \underline{40.80} & 2.67 & 29.33 \\
			DeepSeek-V3.2 & 63.16 & 26.62 & 11.73 & 37.00 & 5.00 & 26.13 \\
			Qwen2.5-32B-Instruct & 63.49 & 29.40 & 16.87 & 40.60 & 1.00 & 26.93 \\
			GLM-4-Plus & 61.84 & 23.84 & 14.11 & 37.00 & 3.33 & 26.73 \\
			LLaMA-3.1-70B-Instruct & 59.21 & 27.31 & 9.21 & 36.00 & 2.33 & 25.53 \\
			Qwen3-30B-A3B-Instruct & 51.64 & 16.44 & 13.30 & 35.80 & 1.33 & 27.67 \\
			LLaMA-3.1-8B-Instruct & 54.28 & 15.97 & 13.26 & 33.20 & 1.33 & 25.93 \\
			Qwen2.5-7B-Instruct & 62.50 & 17.36 & 4.05 & 37.20 & 1.00 & 29.40 \\
			Mistral-7B-Instruct-v0.3 & 63.49 & 18.29 & 4.89 & 30.00 & 1.33 & 25.20 \\
			GLM-4-9B-Chat & 58.55 & 18.52 & 5.44 & 37.60 & 1.00 & 26.07 \\
			\midrule
			\multicolumn{7}{c}{\textit{Uyghur}} \\
			\midrule
			Gemini-3-Flash-Preview & 74.34 & \textbf{\underline{86.34}} & 34.25 & \textbf{\underline{63.00}} & 65.33 & \textbf{\underline{84.73}} \\
			Claude-Sonnet-4.5 & \underline{75.99} & 83.56 & 28.27 & \underline{59.80} & \underline{83.33} & 61.80 \\
			Qwen3-Max & 70.72 & 79.40 & 37.83 & 54.20 & 30.33 & 58.13 \\
			GPT-5.1-Chat & \textbf{\underline{77.63}} & \underline{83.80} & 29.81 & 58.60 & \textbf{\underline{83.67}} & \underline{71.13} \\
			Qwen3-Next-80B-A3B-Instruct & 64.47 & 82.18 & 28.28 & 49.20 & 16.67 & 49.27 \\
			DeepSeek-V3.2 & 61.84 & 75.46 & 35.88 & 45.00 & 32.00 & 39.53 \\
			Qwen2.5-32B-Instruct & 63.49 & 65.05 & 32.81 & 46.20 & 16.00 & 35.67 \\
			GLM-4-Plus & 61.84 & 78.70 & 38.41 & 41.20 & 12.67 & 39.80 \\
			LLaMA-3.1-70B-Instruct & 59.87 & 78.94 & \textbf{\underline{45.88}} & 40.80 & 9.33 & 42.07 \\
			Qwen3-30B-A3B-Instruct & 48.03 & 73.38 & \underline{38.95} & 44.00 & 3.00 & 39.00 \\
			LLaMA-3.1-8B-Instruct & 36.51 & 69.44 & 23.98 & 35.00 & 4.00 & 30.53 \\
			Qwen2.5-7B-Instruct & 64.47 & 59.49 & 30.62 & 42.00 & 6.67 & 31.00 \\
			Mistral-7B-Instruct-v0.3 & 59.54 & 24.77 & 10.44 & 29.80 & 2.33 & 25.40 \\
			GLM-4-9B-Chat & 61.18 & 65.05 & 10.46 & 25.00 & 1.67 & 21.73 \\
			\bottomrule
		\end{tabular}%
	}
	\caption{Performance Comparison of 14 LLMs Across Six Foundation Tasks in Tibetan, Mongolian, and Uyghur. CR: Coreference Resolution; TC: Text Classification; MRC: Machine Reading Comprehension; NLI: Natural Language Inference; MR: Math Reasoning; GDC: General Domain Competence. \textbf{\underline{Bold underline}} indicates first place, \underline{underline} indicates second place.}
	\label{tab:foundation_performance}
\end{table*}

\begin{table*}[htbp]
	\centering
	\scalebox{0.95}{%
		\begin{tabular}{lcccccc}
			\toprule
			\textbf{Model} & \textbf{MCQA} & \textbf{MLIQA} & \textbf{MLE} & \textbf{MLU} & \textbf{MDC} & \textbf{MMT} \\
			\midrule
			\multicolumn{7}{c}{\textit{Tibetan}} \\
			\midrule
			Gemini-3-Flash-Preview & \underline{42.87} & \textbf{\underline{78.66}} & \textbf{\underline{89.43}} & \textbf{\underline{83.00}} & \textbf{\underline{53.13}} & \underline{53.94} \\
			Claude-Sonnet-4.5 & \textbf{\underline{48.20}} & \underline{59.70} & \underline{74.53} & \underline{56.00} & \underline{40.87} & 35.70 \\
			Qwen3-Max & 14.84 & 26.72 & 77.92 & 43.00 & 33.16 & 34.92 \\
			GPT-5.1-Chat & 17.55 & 19.16 & 76.04 & 48.00 & 37.87 & 19.79 \\
			Qwen3-Next-80B-A3B-Instruct & 8.55 & 11.20 & 70.94 & 36.00 & 32.60 & 14.00 \\
			DeepSeek-V3.2 & 29.61 & 32.88 & 66.42 & 26.00 & 28.93 & \textbf{\underline{56.27}} \\
			Qwen2.5-32B-Instruct & 4.90 & 6.13 & 54.91 & 27.00 & 31.40 & 11.08 \\
			GLM-4-Plus & 32.36 & 26.79 & 73.96 & 37.00 & 34.40 & 11.87 \\
			LLaMA-3.1-70B-Instruct & 11.61 & 11.66 & 67.55 & 31.00 & 27.53 & 25.98 \\
			Qwen3-30B-A3B-Instruct & 7.73 & 8.76 & 62.26 & 27.00 & 32.47 & 16.69 \\
			LLaMA-3.1-8B-Instruct & 7.06 & 10.20 & 52.83 & 26.00 & 28.73 & 4.79 \\
			Qwen2.5-7B-Instruct & 2.44 & 5.39 & 36.60 & 30.00 & 24.53 & 6.48 \\
			Mistral-7B-Instruct-v0.3 & 12.65 & 12.95 & 32.83 & 26.00 & 27.53 & 3.56 \\
			GLM-4-9B-Chat & 2.29 & 4.21 & 47.55 & 18.00 & 8.73 & 6.65 \\
			\midrule
			\multicolumn{7}{c}{\textit{Mongolian}} \\
			\midrule
			Gemini-3-Flash-Preview & \textbf{\underline{14.93}} & \textbf{\underline{52.01}} & \textbf{\underline{96.98}} & \textbf{\underline{34.00}} & \textbf{\underline{56.73}} & \textbf{\underline{34.63}} \\
			Claude-Sonnet-4.5 & \underline{15.84} & \underline{23.38} & \underline{77.36} & 22.00 & \underline{33.33} & \underline{21.16} \\
			Qwen3-Max & 6.35 & 8.78 & 84.91 & 24.00 & 30.80 & 14.25 \\
			GPT-5.1-Chat & 1.65 & 0.21 & 44.72 & 21.00 & 24.07 & 4.52 \\
			Qwen3-Next-80B-A3B-Instruct & 1.34 & 3.08 & 56.04 & 21.00 & 29.13 & 10.33 \\
			DeepSeek-V3.2 & 1.03 & 4.23 & 26.42 & 29.00 & 21.40 & 10.89 \\
			Qwen2.5-32B-Instruct & 2.64 & 4.85 & 36.79 & 25.00 & 29.80 & 1.88 \\
			GLM-4-Plus & 0.39 & 2.00 & 30.00 & \underline{30.00} & 25.47 & 5.06 \\
			LLaMA-3.1-70B-Instruct & 4.72 & 3.57 & 31.32 & 28.00 & 21.60 & 7.21 \\
			Qwen3-30B-A3B-Instruct & 2.00 & 1.62 & 30.57 & 26.00 & 26.47 & 1.43 \\
			LLaMA-3.1-8B-Instruct & 1.64 & 8.75 & 29.06 & 27.00 & 21.93 & 2.08 \\
			Qwen2.5-7B-Instruct & 0.31 & 2.40 & 26.60 & 25.00 & 21.80 & 0.13 \\
			Mistral-7B-Instruct-v0.3 & 4.40 & 3.68 & 25.28 & 24.00 & 31.13 & 10.68 \\
			GLM-4-9B-Chat & 1.78 & 3.64 & 20.94 & 28.00 & 12.20 & 0.56 \\
			\midrule
			\multicolumn{7}{c}{\textit{Uyghur}} \\
			\midrule
			Gemini-3-Flash-Preview & \underline{49.45} & \textbf{\underline{93.75}} & \textbf{\underline{97.17}} & 49.00 & \textbf{\underline{61.73}} & \textbf{\underline{45.61}} \\
			Claude-Sonnet-4.5 & \textbf{\underline{54.73}} & \underline{74.46} & \underline{89.81} & 47.00 & 50.40 & 28.71 \\
			Qwen3-Max & 23.62 & 47.26 & 92.64 & \underline{51.00} & 47.73 & 22.82 \\
			GPT-5.1-Chat & 49.14 & 68.37 & 92.83 & \textbf{\underline{56.00}} & \underline{52.27} & 30.51 \\
			Qwen3-Next-80B-A3B-Instruct & 11.77 & 20.63 & 90.57 & 43.00 & 45.00 & 12.24 \\
			DeepSeek-V3.2 & 37.81 & 62.79 & 87.36 & 34.00 & 34.13 & \underline{39.01} \\
			Qwen2.5-32B-Instruct & 2.59 & 4.89 & 84.34 & 41.00 & 37.20 & 10.35 \\
			GLM-4-Plus & 34.32 & 43.50 & 91.13 & 38.00 & 39.67 & 29.10 \\
			LLaMA-3.1-70B-Instruct & 17.38 & 12.72 & 88.87 & 31.00 & 32.53 & 13.78 \\
			Qwen3-30B-A3B-Instruct & 5.93 & 9.19 & 87.36 & 40.00 & 41.27 & 17.18 \\
			LLaMA-3.1-8B-Instruct & 13.15 & 8.62 & 78.11 & 24.00 & 32.40 & 3.17 \\
			Qwen2.5-7B-Instruct & 3.53 & 4.17 & 71.89 & 28.00 & 28.67 & 5.83 \\
			Mistral-7B-Instruct-v0.3 & 3.85 & 10.36 & 37.92 & 30.00 & 23.40 & 3.29 \\
			GLM-4-9B-Chat & 0.55 & 2.70 & 74.34 & 32.00 & 11.33 & 5.67 \\
			\bottomrule
		\end{tabular}%
	}
	\caption{Performance Comparison of 14 LLMs Across Six Chinese Minority Knowledge Tasks in Tibetan, Mongolian, and Uyghur. MCQA: Minority Culture QA; MLIQA: Minority Language Instruction QA; MLE: Minority Language Expressions; MLU: Minority Language Understanding; MDC: Minority Domain Competence; MMT: Minority Machine Translation. \textbf{\underline{Bold underline}} indicates first place, \underline{underline} indicates second place.}
	\label{tab:model_performance}
\end{table*}

\begin{table*}[htbp]
	\centering
	\scalebox{0.95}{%
		\begin{tabular}{lccccc}
			\toprule
			\textbf{Model} & \textbf{CCC} & \textbf{DD} & \textbf{RPE} & \textbf{SSE} & \textbf{VAA} \\
			\midrule
			\multicolumn{6}{c}{\textit{Tibetan}} \\
			\midrule
			Gemini-3-Flash-Preview& 86.61 & \underline{68.13} & \underline{82.95} & 43.30 & \textbf{\underline{92.44}} \\
			Claude-Sonnet-4.5 & 67.72 & \textbf{\underline{69.01}} & 46.31 & \underline{51.55} & \underline{83.12} \\
			Qwen3-Max & 88.98 & 59.56 & 53.41 & 46.39 & 77.08 \\
			GPT-5.1-Chat  & 64.57 & 56.92 & 45.17 & 47.42 & 81.36 \\
			Qwen3-Next-80B-A3B-Instruct & 53.54 & 35.38 & 23.58 & 38.14 & 64.23 \\
			DeepSeek-V3.2 & 51.97 & 41.10 & 33.24 & 39.18 & 65.49 \\
			Qwen2.5-32B-Instruct & \textbf{\underline{94.49}} & 64.62 & \textbf{\underline{71.59}} & \textbf{\underline{84.54}} & 51.13 \\
			GLM-4-Plus & 47.24 & 35.60 & 28.69 & 39.18 & 62.72 \\
			LLaMA-3.1-70B-Instruct & 57.09 & 44.40 & 36.93 & 40.21 & 74.31 \\
			Qwen3-30B-A3B-Instruct & 43.31 & 39.56 & 28.12 & 24.74 & 41.81 \\
			LLaMA-3.1-8B-Instruct & \underline{90.94} & 69.45 & 66.19 & 57.73 & 47.61 \\
			Qwen2.5-7B-Instruct & 38.58 & 37.80 & 16.19 & 31.96 & 43.83 \\
			Mistral-7B-Instruct-v0.3 & 38.19 & 28.13 & 12.22 & 29.90 & 44.08 \\
			GLM-4-9B-Chat & 16.54 & 40.00 & 9.66 & 14.43 & 24.94 \\
			\midrule
			\multicolumn{6}{c}{\textit{Mongolian}} \\
			\midrule
			Gemini-3-Flash-Preview & 86.61 & 56.70 & \underline{79.26} & 47.42 & \textbf{\underline{85.14}} \\
			Claude-Sonnet-4.5 & 64.57 & 49.45 & 39.77 & \underline{57.73} & \underline{49.37} \\
			Qwen3-Max & 67.32 & 54.73 & 62.78 & \underline{57.73} & 47.61 \\
			GPT-5.1-Chat & 82.68 & \underline{64.62} & 65.06 & 62.89 & 44.84 \\
			Qwen3-Next-80B-A3B-Instruct & 63.78 & 53.19 & 36.08 & 54.64 & 45.09 \\
			DeepSeek-V3.2 & 54.72 & 52.31 & 47.73 & 43.30 & 40.55 \\
			Qwen2.5-32B-Instruct & \textbf{\underline{91.34}} & \textbf{\underline{68.57}} & \textbf{\underline{72.16}} & \textbf{\underline{67.01}} & 48.36 \\
			GLM-4-Plus & 30.71 & 38.90 & 26.42 & 46.39 & 35.01 \\
			LLaMA-3.1-70B-Instruct  & 32.68 & 35.38 & 30.68 & 38.14 & 30.23 \\
			Qwen3-30B-A3B-Instruct & 58.27 & 55.60 & 49.72 & \textbf{\underline{67.01}} & 39.55 \\
			LLaMA-3.1-8B-Instruct & \underline{74.80} & 46.15 & 46.31 & 56.70 & 35.52 \\
			Qwen2.5-7B-Instruct & 7.09 & 25.93 & 12.22 & 20.62 & 27.96 \\
			Mistral-7B-Instruct-v0.3 & 5.12 & 23.30 & 9.09 & 31.96 & 24.69 \\
			GLM-4-9B-Chat & 48.43 & 47.47 & 51.42 & 31.96 & 29.72 \\
			\midrule
			\multicolumn{6}{c}{\textit{Uyghur}} \\
			\midrule
			Gemini-3-Flash-Preview & 87.40 & \underline{82.42} & \underline{73.58} & \underline{83.51} & \textbf{\underline{86.15}} \\
			Claude-Sonnet-4.5 & 77.17 & 76.48 & 57.67 & 68.04 & \underline{69.52} \\
			Qwen3-Max & 70.87 & 71.21 & 52.56 & 61.86 & 63.48 \\
			GPT-5.1-Chat & \underline{90.55} & \textbf{\underline{84.62}} & \textbf{\underline{76.70}} & \textbf{\underline{79.38}} & 71.79 \\
			Qwen3-Next-80B-A3B-Instruct & 72.05 & 72.53 & 57.67 & 64.95 & 65.49 \\
			DeepSeek-V3.2 & 61.02 & 60.66 & 48.58 & 50.52 & 39.80 \\
			Qwen2.5-32B-Instruct & 35.04 & 48.35 & 39.49 & 34.02 & 32.49 \\
			GLM-4-Plus & 80.31 & 56.70 & 42.61 & 32.99 & 53.40 \\
			LLaMA-3.1-70B-Instruct & 38.98 & 55.82 & 42.90 & 26.80 & 42.07 \\
			Qwen3-30B-A3B-Instruct & 76.77 & 53.63 & 43.18 & 44.33 & 50.13 \\
			LLaMA-3.1-8B-Instruct & 57.87 & 53.63 & 37.22 & 43.30 & 38.54 \\
			Qwen2.5-7B-Instruct & 74.02 & 55.38 & 54.83 & 76.29 & 63.73 \\
			Mistral-7B-Instruct-v0.3 & \textbf{\underline{91.34}} & 51.43 & 67.05 & 60.82 & 72.29 \\
			GLM-4-9B-Chat & 14.96 & 14.07 & 6.82 & 0.00 & 13.10 \\
			\bottomrule
		\end{tabular}%
	}
	\caption{Performance Comparison of 14 LLMs Across Five Safety Alignment Tasks in Tibetan, Mongolian, and Uyghur. CCC: Commercial Compliance Check; DD: Discrimination Detection; RPE: Rights Protection Evaluation; SSE: Service Safety Evaluation; VAA: Value Alignment Assessment. \textbf{\underline{Bold underline}} indicates first place, \underline{underline} indicates second place.}
	\label{tab:safety_performance}
\end{table*}

\begin{table*}[htbp]
	\centering
	\caption{Summary of Evaluated LLMs}
	\label{model_details}
	\begin{tabular}{cc}
		\toprule
		\textbf{Model (\#params)} & \textbf{Version} \\
		\midrule
		\multicolumn{2}{c}{\textit{\textbf{Commercial Models}}} \\
		\midrule
		GPT-5.1-Chat & gpt-5.1-chat \\
		Claude-Sonnet-4.5 & claude-sonnet-4-5-20250929 \\
		Gemini-3-Flash-Preview & google\_gemini-3-flash-preview \\
		Qwen3-Max & qwen\_qwen3-max \\
		GLM-4-Plus & glm-4-plus \\
		DeepSeek-V3.2 (685B) & deepseek\_deepseek-v3.2 \\
		\midrule
		\multicolumn{2}{c}{\textit{\textbf{Open-Source Models}}} \\
		\midrule
		Mistral-7B-Instruct-v0.3 (7B) & Mistral-7B-Instruct-v0.3 \\
		Qwen2.5-7B-Instruct (7B) & Qwen2.5-7B-Instruct \\
		LLaMA-3.1-8B-Instruct (8B) & Meta-LLaMA-3.1-8B-Instruct \\
		GLM-4-9B-Chat (9B) & glm-4-9b-chat \\
		Qwen3-30B-A3B-Instruct (30B) & qwen\_qwen3-30b-a3b-instruct-2507 \\
		Qwen2.5-32B-Instruct (32B) & Qwen2.5-32B-Instruct \\
		LLaMA-3.1-70B-Instruct (70B) & meta-llama\_llama-3.1-70b-instruct \\
		Qwen3-Next-80B-A3B-Instruct (80B) & qwen\_qwen3-next-80b-a3b-instruct \\
		\bottomrule
	\end{tabular}
\end{table*}

\begin{table*}[htbp]
	\centering
	\caption{Task Details Per Language in \textsc{CMiLBench}}
	\label{task_details}
	\begin{tabular}{ccccc}
		\toprule
		\makecell{\textbf{Task}\\\textbf{Category}} & \textbf{Task Name} & \textbf{Per Language} & \textbf{Task Type} & \makecell{\textbf{Evaluation}\\\textbf{Metric}} \\
		\midrule
		\multirow{6}{*}{\makecell{\small{Foundation}\\\small{Tasks}}}
		& \small{Natural Language Inference} & \small{500} & \small{Multiple Choice} & \small{Accuracy} \\
		& \small{Coreference Resolution} & \small{304} & \small{Multiple Choice} & \small{Accuracy} \\
		& \small{Machine Reading Comprehension} & \small{500} & \small{Generation} & \small{ROUGE-L} \\
		& \small{Text Classification} & \small{432} & \small{Fill in the Blank} & \small{Accuracy} \\
		& \small{Math Reasoning} & \small{300} & \small{Fill in the Blank} & \small{Accuracy} \\
		& \small{General Domain Competence} & \small{1,500} & \small{Multiple Choice} & \small{Accuracy} \\
		\midrule
		\multirow{6}{*}{\makecell{\small{Chinese}\\\small{Minority}\\\small{Knowledge}\\\small{Tasks}}} 
		& \small{Minority Language Expressions} & \small{530} & \small{Multiple Choice} & \small{Accuracy} \\
		& \small{Minority Machine Translation} & \small{500} & \small{Generation} & \small{BLEU \& chrF++} \\
		& \small{Minority Culture QA} & \small{282} & \small{Generation} & \small{LLM-as-a-Judge} \\
		& \small{Minority Language Instruction QA} & \small{218} & \small{Generation} & \small{LLM-as-a-Judge} \\
		& \small{Minority Language Understanding} & \small{100} & \small{Multiple Choice} & \small{Accuracy} \\
		& \small{Minority Domain Competence} & \small{1,500} & \small{Multiple Choice} & \small{Accuracy} \\
		\midrule
		\multirow{5}{*}{\makecell{\small{Safety}\\\small{Alignment}\\\small{Tasks}}} 
		& \small{Discrimination Detection} & \small{455} & \small{Multiple Choice} & \small{Accuracy} \\
		& \small{Value Alignment Assessment} & \small{397} & \small{Multiple Choice} & \small{Accuracy} \\
		& \small{Rights Protection Evaluation} & \small{352} & \small{Multiple Choice} & \small{Accuracy} \\
		& \small{Commercial Compliance Check} & \small{254} & \small{Multiple Choice} & \small{Accuracy} \\
		& \small{Service Safety Evaluation} & \small{97} & \small{Multiple Choice} & \small{Accuracy} \\
		\midrule
		\textbf{Total} & \textbf{17 tasks} & \textbf{8,221 samples} & \textbf{3 types} & \textbf{5 metrics} \\
		\bottomrule
	\end{tabular}
\end{table*}

\begin{table*}[htbp]
	\centering
	\caption{Multi-Dimensional Scoring Framework for LLM-as-a-Judge Evaluation}
	\label{scoring_dimensions}
	\begin{tabular}{p{3.5cm} l p{7.7cm}}
		\toprule
		\textbf{Task Name} & \textbf{Response Type} & \textbf{Scoring Dimensions} \\
		\midrule
		\textbf{Minority Culture QA} & -- & Factual Accuracy, Cultural Understanding Depth, Appropriateness of Language Use, Content Completeness, Authenticity of Insider Perspective \\
		\midrule
		\multirow{8}{=}{\textbf{Minority Language Instruction QA}} 
		& Factual Answer & Factual Accuracy, User Need Satisfaction, Clarity, Completeness \\
		\cmidrule(l){2-3}
		& Reasoning Answer & Factual Accuracy, User Need Satisfaction, Logical Coherence, Completeness \\
		\cmidrule(l){2-3}
		& Generation Answer & Factual Accuracy, User Need Satisfaction, Logical Coherence, Creativity, Richness \\
		\cmidrule(l){2-3}
		& Advisory Answer & Factual Accuracy, User Need Satisfaction, Fairness \& Responsibility, Creativity \\
		\bottomrule
	\end{tabular}
\end{table*}

\begin{table*}[htbp]
	\centering
	\caption{Comparison between \textit{Minority Culture QA} and \textit{Minority Domain Competence}}
	\label{task_comparison_1}
	\renewcommand{\arraystretch}{1.5}
	\begin{tabular}{l p{6cm} p{6cm}}
		\toprule
		\textbf{Feature} & \textbf{Minority Culture QA} & \textbf{Minority Domain Competence} \\
		\midrule
		\textbf{Content Type} & Sociological, Anthropological, \& Cultural & Technical, Scientific, \& Academic History \\
		\textbf{Format} & Open-ended Text (Descriptive) & Multiple Choice (Objective) \\
		\textbf{Example Topics} & Marriage Customs, Farming Tools & Medical Prescriptions, Specific Historical Regimes \\
		\textbf{Primary Goal} & Explain Cultural Context and Traditions & Test Specific Professional Competency \\
		\bottomrule
	\end{tabular}
\end{table*}

\begin{table*}[htbp]
	\centering
	\caption{Comparison between \textit{Minority Language Understanding} and \textit{Minority Language Expressions}}
	\label{task_comparison_2}
	\renewcommand{\arraystretch}{1.5}
	\begin{tabular}{l p{6cm} p{6cm}}
		\toprule
		\textbf{Feature} & \textbf{Minority Language Understanding} & \textbf{Minority Language Expressions} \\
		\midrule
		\textbf{Primary Task} & Grammar, Logic, \& Reading Comprehension & Translation \& Vocabulary Mapping \\
		\textbf{Language Direction} & Monolingual (Tibetan Question $\rightarrow$ Tibetan Answer) & Bilingual (Chinese Question $\rightarrow$ Tibetan Answer) \\
		\textbf{Key Skills} & Syntax, Particle Usage, Logical Reasoning & Idiomatic Translation, Lexical Knowledge \\
		\bottomrule
	\end{tabular}
\end{table*}

\end{document}